\documentclass{article}

\usepackage{arxiv}

\usepackage[utf8]{inputenc} 
\usepackage[T1]{fontenc}    
\usepackage{url}            
\usepackage{booktabs}       
\usepackage{amsfonts}       
\usepackage{nicefrac}       
\usepackage{microtype}      
\usepackage{xcolor}         
\usepackage{bm}
\usepackage{array}
\usepackage[utf8]{inputenc} 
\usepackage[T1]{fontenc}    
\usepackage{epsfig}
\usepackage{epstopdf}
\usepackage{graphics}
\usepackage{graphicx}
\usepackage[normalem]{ulem}
\usepackage{float}
\usepackage{colortbl}
\usepackage{subfigure}
\usepackage{multirow}
\usepackage{diagbox}
\usepackage{bbding}
\usepackage{mathrsfs}
\usepackage{amsmath} 
\usepackage{amsthm}
\usepackage{longtable}
\usepackage{makecell}
\usepackage{enumitem}
\usepackage{color}
\usepackage{algorithm}
\usepackage{algorithmic}
\usepackage{caption}

\newcolumntype{L}[1]{>{\raggedright\let\newline\\\arraybackslash\hspace{0pt}}m{#1}}
\newcolumntype{C}[1]{>{\centering\let\newline  \\\arraybackslash\hspace{0pt}}m{#1}}
\newcolumntype{R}[1]{>{\raggedleft\let\newline \\\arraybackslash\hspace{0pt}}m{#1}}

\newtheorem{thm}{Theorem}[section]

\newtheorem{prop}{Proposition}[section]

\newtheorem{defn}{Definition}[section]

\newtheorem{remark}{Remark}

\title{Automorphic Equivalence-aware Graph Neural Network}

\author{%
  Fengli Xu$^{1}$, 
  \quad 
  Quanming Yao$^{1,2}$, 
  \quad
  Pan Hui$^{3}$, 
  \quad
  Yong Li$^{1}$
  \\
  $^{1}$BNRIST \& EE Department, Tsinghua University\\
  $^{2}$4Paradigm Inc. \\
  $^{3}$CSE Department, HKUST. \\
  \texttt{liyong07@tsinghua.edu.cn}\\
}

\begin{document}

\maketitle

\begin{abstract}
Distinguishing the automorphic equivalence of nodes in a graph plays an essential role in many scientific domains, \emph{e.g.}, computational biologist and social network analysis. However, existing graph neural networks (GNNs) fail to capture such an important property. To make GNN aware of automorphic equivalence, we first introduce a localized variant of this concept --- ego-centered automorphic equivalence~(Ego-AE). Then, we design a novel variant of GNN, \emph{i.e.}, GRAPE, that uses learnable AE-aware aggregators to explicitly differentiate the Ego-AE of each node's neighbors with the aids of various subgraph templates. While the design of subgraph templates can be hard, we further propose a genetic algorithm to automatically search them from graph data. Moreover, we theoretically prove that GRAPE is expressive in terms of generating distinct representations for nodes with different Ego-AE features, which fills in a fundamental gap of existing GNN variants. Finally, we empirically validate our model on eight real-world graph data, including social network, e-commerce co-purchase network, and citation network, and show that it consistently outperforms existing GNNs. The source code is public available at~\url{https://github.com/tsinghua-fib-lab/GRAPE}.
\end{abstract}

\section{Introduction}
\label{sec:intro}

The past few years have witnessed the phenomenal success of GNNs in numerous  
graph learning tasks, such as node classification~\cite{kipf2016semi}, 
link prediction~\cite{you2019position}, 
and community detection~\cite{hamilton2017inductive}, 
which is largely due to their capability of simultaneously modelling the connecting patterns and feature distribution in each node's local neighborhood. 
As a result, it leads to a surge of interests from both academia and industry to develop more powerful GNN models~\cite{wu2020comprehensive}. 
Despite of various architectures, the most popular GNNs, like GCN~\cite{kipf2016semi}, GraphSAGE~\cite{hamilton2017inductive}, 
GAT~\cite{velivckovic2017graph} and GIN~\cite{xu2018powerful}, 
apply permutation invariant aggregate function on each node's local neighborhood to learn node embeddings, 
which leads to concerns about their representational power~\cite{garg2020generalization,xu2018powerful}.

In this paper, 
we investigate GNN's expressiveness from an important but largely overlooked angle, 
\emph{i.e.}, the capacity to distinguish automorphic equivalence within each node's local neighborhood.
Automorphic equivalence (AE)~\cite{everett1985role} is a classic concept to differentiate the structural role of each node in a given graph. 
Specifically, two nodes are considered to be AE only if they are interchangeable in some index permutations that preserve the connection matrix, \emph{i.e.}, 
graph automorphisms~\cite{mckay2014practical}. AE can identify the nodes that exhibit identical structural features in a graph, which makes it a central topic in computational biologist, social network analysis and other scientific domains~\cite{lorrain1971structural,luczkovich2003defining}. For example, empirical studies show AE is an important indicator of social position and behavior similarity in social network~\cite{friedkin1997social, mizruchi1993cohesion}, which thus might significantly benefit GNN architecture design.

Empirically efficient heuristics have been proposed to identify the AE in moderate scale graphs by enumerating all the possible automorphisms~\cite{colbourn1981linear, sparrow1993linear}.
However, previous analytic methods only classify nodes into categorical equivalence sets. 
Although categorical features can be jointly optimized with GNNs under various frameworks~\cite{liu2021graph,xie2021self}, 
little previous efforts are invested to principally incorporate AE into GNNs. 
Thus, we aim to design a novel GNN model that is provably expressive in capturing AE features and can be tuned in a data-dependent manner based on the graph data and targeted applications, 
which will effectively allow us to learn expressive function to harness the power of AE feature.

Here, we propose \underline{GR}aph \underline{A}utomor\underline{P}hic \underline{E}quivalent network, \emph{i.e.}, GRAPE, a novel variant of GNN that can learn expressive representation by differentiating the automorphic equivalences of each node's neighbors. First, GRAPE extends the classic AE concept into a localized setting, \emph{i.e.}, Ego-AE, to accommodate the local nature of GNNs. Specifically, Ego-AE identifies the local neighborhoods of each node by mapping with given subgraph templates and then partitions the neighboring nodes into Ego-AE sets based on the graph automorphisms in neighborhood. Second, we design a learnable AE-aware aggregators to model the node features in these Ego-AE sets, which adaptively assigns different weights to neighboring nodes in different Ego-AE sets and explicitly model the interdependency among them. Moreover, in order to capture complex structural features, GRAPE proposes to fuse the embeddings learned from Ego-AE sets identified by different subgraph templates with a squeeze-and-excitation module~\cite{hu2018squeeze}. Finally, to alleviate the barrier of subgraph template design, we propose an efficient genetic algorithm to automatically search for optimal subgraph templates. Specifically, it gradually optimizes a randomly initiated population of subgraph templates by iteratively exploring the adjacency of good performing candidates and eliminating the bad performing ones. 
To accelerate the search process, we further design an incremental subgraph matching algorithm that can leverage the similarity between subgraphs to greatly reduce the complexity of finding matched instances.   

We theoretically prove that the proposed GRAPE is expressive in terms of learning distinct representations for nodes with different Ego-AE sets, which fundamentally makes up the shortcomings of popular GNN variants, \emph{e.g.}, GCN~\cite{kipf2016semi}, GraphSAGE~\cite{hamilton2017inductive}, GAT~\cite{velivckovic2017graph} and GIN~\cite{xu2018powerful}. 
Moreover, we empirically validate GRAPE on eight real-world datasets, which cover the scenarios of social network, citation network and e-commerce co-purchase network. 
Experiments show GRAPE is consistently the best performing model across all datasets with up to 26.7\% accuracy improvement.
Besides, case studies indicate GRAPE can effectively differentiate the structural roles of each node's neighbors. Moreover, the proposed genetic algorithm efficiently generates high quality subgraph templates that have comparable performance with the hand-crafted ones. 

\section{Related Works}
\label{sec:rel:MPNN}

In the sequel,
we define a graph as $G = (\mathcal{V}, \mathcal{E})$, 
where $\mathcal{V} = \{v_1$,$\cdots$,$v_n\}$ is the set of nodes and $\mathcal{E}=\{(v_i, v_j)\}$ is the set of edges. 
Let the feature vector of node $v_i$ be $\mathcal{X}(v_i)$, 
and $\mathcal{N}(v_i)$ represents the set of $v_i$'s neighbor nodes. 
$N$ is the number of node and $M$ is the embedding size.

\subsection{Automorphic Equivalence (AE) in Graph Analysis}

Here, we investigate one of the most popular structural equivalence concepts, 
\emph{i.e.,} automorphic equivalence (AE)~\cite{everett1985role}
which plays a central role in 
computational biologist, social network analysis and other scientific domains~\cite{lorrain1971structural,luczkovich2003defining}.
The most interesting part of AE is that it identifies the nodes with exact same structural patterns, \emph{e.g.}, degree and centrality~\cite{everett1990ego}, 
but not necessarily connecting to the same neighboring nodes.
Basically,
AE is defined as follows. 

\begin{defn} 
	\label{def:equivalence}
	Given a graph $G = (\mathcal{V}, \mathcal{E})$, an \textbf{automporphism} $\pi(\star)$ is a node permutation that preserves the adjacency matrix, 
	i.e., the permuted nodes $\pi(v_{a})$ and $\pi(v_{b})$ are connected if and only if nodes $v_{a}$ and $v_{b}$ are connected.
	Two nodes $v_{a}$ and $v_{b}$ are considered to be \textbf{\underline{A}utomorphic \underline{E}quivalence (AE)} if there is a graph automorphism that maps one onto the other, 
	i.e., $\pi(v_{a}) = v_{b}$.
\end{defn} 

\begin{figure}[t]
	\centering
	\includegraphics[width=1.0\textwidth]{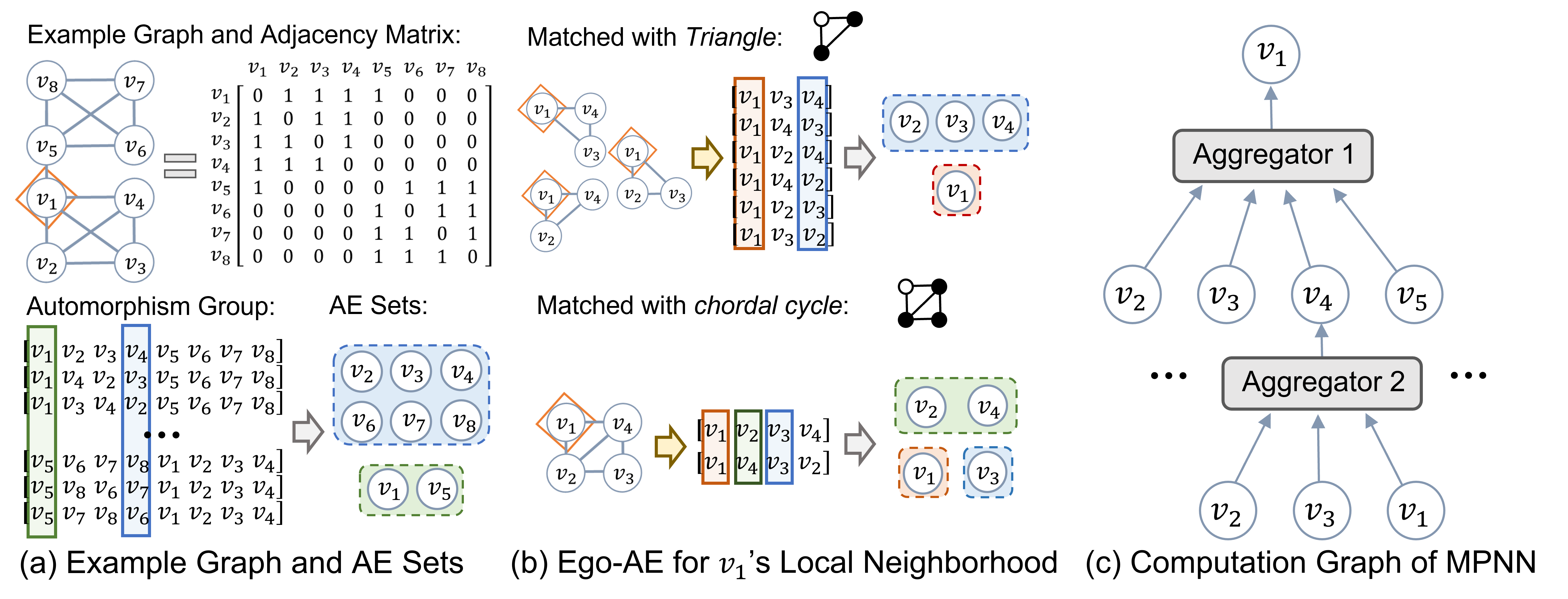}
	\caption{Illustration of AE and the limitation of MPNN framework.}
	\label{fig:motive}
\end{figure}

Nodes in $\mathcal{V}$ that
are AE with each other constitutes a AE set. An example of such sets is in Figure~\ref{fig:motive}~(a). 
Specifically, AE sets can be identified by enumerating the automorphism group of the given graph with efficient \emph{Nauty} algorithm~\cite{mckay2014practical}, 
and the nodes that are mapped onto each others in different automorphisms will be partitioned into same AE sets.

Previous works have attempted to preserve the structural similarities on graph via various node embedding algorithms~\cite{donnat2018learning,ahmed2020role,ribeiro2017struc2vec}. 
Recently, GraphWave~\cite{donnat2018learning} was proposed to leverage wavelet diffusion patterns to capture the structural roles in node representations. 
Besides, role2vec~\cite{ahmed2020role} introduced
a generalized feature-based random walks that aims to represent the structural similarities among nodes.

However, the connection between AE and GNN has not been examined in existing literature. Moreover, AE is defined on whole graph level, which is infeasible to compute in large graphs and goes against the inherent local nature of most GNN frameworks. In this paper, we aim to extend the AE concept to local setting, 
and propose a novel GNN model to harness its power. 

\subsection{The Expressive Power of GNNs}
The recent success of GNNs draws increasing interests in investigating their capability in capturing structural properties~\cite{xu2018powerful,garg2020generalization}. Specifically, message passing neural network (MPNN) is the most popular GNN framework~\cite{gilmer2017neural,hamilton2017inductive,xu2018powerful}. 

Let $\textsf{AGG}$ be the \emph{aggregate} function that collects feature from neighbors, 
and $\textsf{COMB}$ be the \emph{combine} function that integrates each node's self feature with those from $\textsf{AGG}$. 
Generally, MPNN generates representation for a node $v_t$ at $k$-th layer as
\begin{align}
& \bm{h}^k_{v_t} 
= \textsf{COMB}
\big( 
\bm{h}^{k-1}_{v_t}, \;
\textsf{AGG}
( 
\mathcal{H}_{v_t}^{k - 1}
) 
\big),
\label{eq:mpnn}
\end{align}
where $\bm{h}^0(v_i) = \mathcal{X}(v_i)$,
and $\mathcal{H}_{v_t}^{k - 1}
= \big\{ \bm{h}^{k-1}_{v_j}| v_j \in \mathcal{N}(v_t) \big\}$.
Existing MPNNs use local permutation invariant \textsf{AGG}s to compute node embeddings, 
which subsumes a large class of popular GNN models such as GCN~\cite{kipf2016semi}, GraphSAGE~\cite{hamilton2017inductive}, GAT~\cite{velivckovic2017graph}, Geniepath~\cite{liu2019geniepath} and GIN~\cite{xu2018powerful}. 
The family of MPNNs is proven to be theoretically linked to Weisfeiler-Lehman (WL) subtree kernel~\cite{hamilton2017inductive}.
Subsequently, they are at most as powerful as $1$-WL test on discriminating graph isomorphisms~\cite{xu2018powerful}.

More recently, several attempts have been made to improve the expressiveness of GNN beyond MPNN framework. 
They can mainly be classified into two categories: augmenting node features and designing more powerful architectures. In terms of augmenting node feature, recent works proposed to introduce various additional feature~\cite{sato2019approximation,klicpera2020directional,you2019position}. 3D-GCN uses additional 3D point cloud feature to differentiate neighbors and facilitate learnable graph kernels~\cite{lin2021learning}. 
Moreover, GNN variants can in theory achieve universal approximation on graph by equipping nodes with randomly initialized feature vector~\cite{sato2021random}, but they are difficult to generalize to different graphs in practice~\cite{bouritsas2020improving}. Previous work also proposed to augment node feature with substructure count~\cite{bouritsas2020improving}, which however cannot reveal the local structural roles in each node's neighborhood.
On the other hand, the previous efforts of designing more powerful GNN architecture are dedicated to different structural properties, \emph{e.g.}, graph isomorphism~\cite{xu2018powerful} and graph moment~\cite{dehmamy2019understanding}. Although AE is an important concept for graph data analysis, it has not been addressed by previous GNN research. 
More detailed comparison with the existing GNN variants is provided in Appendix~\ref{app:relate}.

However, previous works have shown that identifying automorphic equivalence is a strictly more difficult task than discriminating graph isomorphisms~\cite{toran2004hardness}. 
Specifically, two AE nodes always have isomorphic neighborhood, 
while the nodes with isomorphic neighborhood are not necessary AE~\cite{everett1990ego}.
Therefore, it raises concerns about MPNN's expressive power of AE feature, which has not been adequately investigated in previous research. In this paper, we aim to design a novel GNN model that is provably expressive in modeling Ego-AE, 
which falling in the category of designing novel GNN architecture. To the best of our knowledge, 
we are the first to empower GNN with the capability of capturing automorphic equivalence.

\subsection{Genetic Algorithm}

Genetic algorithm~\cite{eiben2003introduction} is a widely adopted algorithm
for combinatorial optimization problems,
\emph{e.g.}, traveling salesman problem. 
Recently,
it has also been used to tune hyper-parameters~\cite{lorenzo2017particle} and search neural architectures for deep networks~\cite{yao1999evolving,xie2017genetic}.
Basically,
genetic algorithm
mimics the natural selection process to iteratively to search for better solutions by exploring the adjacency of promising candidates and eliminating the worst-performing ones~\cite{darwin1909origin}. 
Therefore, 
it can iteratively optimize the candidate population from \emph{parent generation} to \emph{children generation}. 
Specifically, genetic algorithms are often made up 
the following four components:
(i) \textit{Mutation}: explores the adjacency of promising candidates in \emph{parent generation} by generating slightly different candidates in \emph{children generation}; 
(ii) \textit{Crossover}: search for different combinations of genetic features in the candidates in \emph{parent generation};
(iii) \textit{Evaluation}: measure the fitness of candidates in given tasks;
(iv) \textit{Selection}: eliminating the candidates with worst performance.

Here, we design a genetic algorithm to automatically optimize the subgraph templates in the proposed GRAPE. It effectively allows us to search the architecture of GRAPE in a data-dependent manner, which significantly reduces the barrier of subgraph template design. 

\begin{figure}[t]
	\centering
	\includegraphics[width=1.0\textwidth]{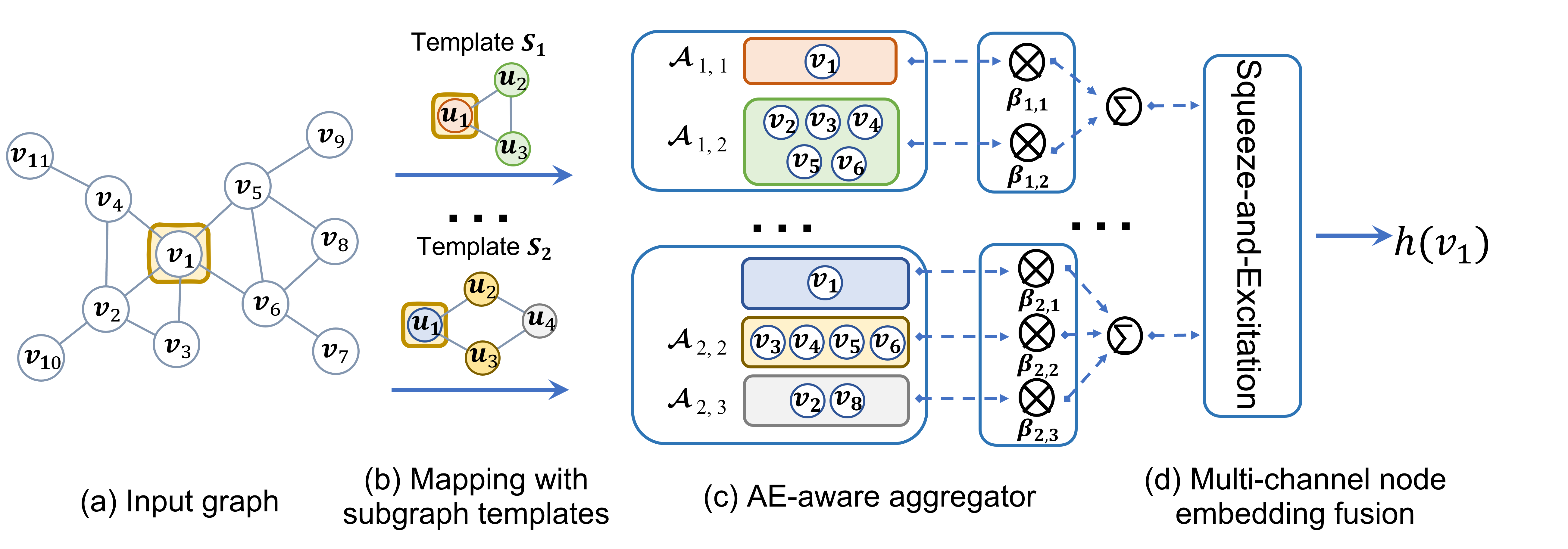}
	\caption{The GRAPE model. (a) The input graph with node $v_1$ as ego node.
		(b) Mapping node $v_1$'s neighborhood with given subgraph templates, where nodes with same color are Ego-AE. 
		(c) Aggregating features from AE sets with learnable AE-aware aggregators. 
		(d) Fusing multi-channel node embedding with squeeze-and-excitation module.}
	\label{fig:model}
\end{figure}

\section{The Proposed Method}
\label{sec:Method}

One prominent feature of most GNN variants is that the node embedding are generated based on the local neighborhood, which significantly improved the scalability and generalization of GNN models. To accommodate the local nature of GNNs, the concept of AE needs to be fundamentally extended and redefined on each node's local neighborhood.
Besides, the local neighborhoods defined by different subgraph templates may exert different influence on the ego node~\cite{ying2019gnn,ahmed2020role}. 
Thus, we propose a subgraph template-dependent local version of AE, \emph{i.e.}, ego-centered automorphic equivalence (Ego-AE). 
Specifically, a subgraph template is defined as a connected graphlet $S = (\mathcal{U}, \mathcal{R})$, 
where $\mathcal{U}$ and $\mathcal{R}$ are the sets of nodes and edges, respectively. 
To differentiate the unique role of ego node, 
we set an anchoring node in subgraph template that always maps to the ego node. Given a graph $G=(\mathcal{V}, \mathcal{E})$, 
a subgraph template $S=(\mathcal{U}, \mathcal{R})$ and a node $v_t$, the Ego-AE on $v_t$'s local neighborhood is defined as follow.

\begin{defn} 
\label{def:ego_ae}
We define $\mathcal{M}_S(v_t)$ as the set of subgraphs that match the subgraph template $S$ in $v_t$'s local neighborhood.
An ego-centered automorphism $\pi_e(\star)$ is an automorphism on the matched subgraphs $m \in \mathcal{M}_S(v_t)$ that has a fixed index of node $v_t$, 
\emph{i.e.}, $\pi_e(v_t)\equiv v_t$.
Two nodes $v_{a}$ and $v_{b}$ are considered to be \textbf{\underline{Ego}-centered \underline{A}utomorphic \underline{E}quivalence (Ego-AE)} 
if there exists an automorphism $\pi_e(\star)$ that maps one onto the other, \emph{i.e.}, $\pi_e(v_{a}) = v_{b}$.
\end{defn}

Without loss of generality, various forms of subgraph template $S$ can be adopted to capture the structural patterns of different semantics.
Figure~\ref{fig:motive}~(b) shows the Ego-AE on $v_1$'s local neighborhoods with the subgraph templates of \emph{triangle} and \emph{chordal cycle}. 
Specifically, we first identify all the matched subgraphs with the anchoring nodes (white color) in subgraph templates fixed to the ego node $v_1$, 
and then the nodes covered by the matched instances are partitioned into Ego-AE sets based on the corresponding automorphisms. 
	We can observe that Ego-AE successfully differentiates the roles of $v_1$'s neighboring nodes based on the structural features.

However, GNN's capacity in capturing Ego-AE is largely unknown in the literature. In fact, we prove that the standard MPNN has fundamental limitations (see Section~\ref{sec:theory}), while other GNN variants focus on different graph properties which largely overlooked Ego-AE. 

\subsection{The Proposed GRAPE Model}
\label{sec:gnnarch}

We aim to propose a novel GNN model, \emph{i.e.}, \underline{GR}aph \underline{A}utomor\underline{P}hic \underline{E}quivalence network (GRAPE), 
that is provably expressive in capturing Ego-AE. 
The overall framework is in Figure~\ref{fig:model}. 
In the sequel, we describe them in details,
and the complete algorithm of GRAPE is in Appendix~\ref{app:grape}.

\subsubsection{AE-aware Aggregator}
Here, we design a novel AE-aware aggregators to learn from the Ego-AE sets in each node's local neighborhood with given subgraph templates.
Specifically, we denote $v_t$'s Ego-AE sets with subgraph template $S_l$ as 
$\mathcal{T}_l=\{\mathcal{A}_{l,1}(v_t), ..., \mathcal{A}_{l,j}(v_t), ..., \mathcal{A}_{l,m_l}(v_t)\}$, 
where $\mathcal{A}_{l,j}(v_t)$ is the set of nodes corresponding to the $j$-th sets of Ego-AE nodes in $S_l$ and $m_l$ is the total number Ego-AE sets.
Then, $v_t$'s node embedding $\bm{h}_l^k(v_t)$ can be computed as follows:
\begin{equation}
\label{eq:gae}
\bm{h}^k_l(v)
= \textsf{MLP}
\big( 
\sum\nolimits_j \beta_{l,j} \cdot 
\sum\nolimits_{v_n \in \mathcal{A}_{l,j}(v_t)} 
\bm{h}_l^{k-1}(v_n)
\big),
\end{equation}
where $\beta_{l,j}$'s are learnable weights that model the importance of $\mathcal{A}_{l,j}(v_t)$ and $\textsf{MLP}(\cdot)$ is a multi-layer perception (MLP) function~\cite{hornik1989multilayer} 
that generates output embeddings. 

Equation~\eqref{eq:gae} is illustrated in Figure~\ref{fig:model}~(c). 
Differs from  MPNNs, the proposed AE-aware aggregator can explicitly differentiate the neighboring nodes with different structural roles by assigning different weights $\beta_{l,j}$ to them. 
It allows GRAPE to capture the combination of important Ego-AE sets and effectively models the interdenpendency among them. 
Note that GRAPE does not have an explicit \emph{COMBINE} function to account for the ego node's self feature, 
since the ego node $v_t$ will always be captured in a unique Ego-AE set, 
such as the $\mathcal{A}_{1,1}$ and $\mathcal{A}_{2,1}$ in Figure~\ref{fig:model}~(c).

\subsubsection{Fusing Embeddings from Different Aggregators}

To simultaneously capture different structural feature, we design a squeeze-and-excitation module to fuse the node embeddings learned from a set of subgraph templates, which is inspired by the channel-wise enhancement technique recently proposed in~\cite{hu2018squeeze}. Specifically, by leveraging a set of subgraph templates $\Omega=\{S_1, S_2, ..., S_L\}$, GRAPE can learn multiple AE-aware aggregators with each subgraph template to capture different structural features respectively.
The GRAPE can learn to assign different weights $\bm{\alpha}^k$ for $\bm{h}_l^k(v_t)$ and generate the fused embedding for $v_t$ as following
\begin{align}
\bm{h}^k(v)
= \sum\nolimits_{l\in 1, ..., L} \bm{\alpha}^k[l]\cdot \bm{h}^k_l(v).
\label{eq:fusing}
\end{align}
Here, the learnable weights $\bm{\alpha}^k$ is computed as following
\begin{align}
\bm{\gamma}^k[l]
= \frac{1}{N}
\sum\nolimits_{n = 1}^N
\textsf{MEAN}(\bm{h}_l^k(v_n) ),
\quad
\bm{\alpha}^k
= \textsf{ReLU}\big(\bm{W}^k_2\cdot\textsf{ReLU}\big(\bm{W}^k_1\cdot\bm{\gamma}^k\big)\big),
\label{eq:alpha}
\end{align}
where $\bm{\gamma}^k[l]$ is the global average pooling on the node embeddings learned with subgraph template $S_l$, 
$\bm{W}^k_1$ and $\bm{W}^k_2$ are two learnable matrices with $\mathbb{R}^{L\times L}$ size,
and \textsf{ReLU} is the relu activation function.

\subsection{Genetic Search of Subgraph Templates}
\label{sec:evoae}

To reduce the barrier of hand-crafted subgraph templates, we formulate the automatic subgraph template design problem as an optimization problem that aims to search for the best performing combinations of subgraph templates $\Omega$. Let the designed GRAPE in Section~\ref{sec:gnnarch} be $F$ with model parameter $\Theta$, which leverages the Ego-AE sets $\{\mathcal{T}_{1}, ..., \mathcal{T}_{L}\}$ identified by \emph{Matching} $\Omega$ on the given graph $G$. 
This subsequently leads to the following bi-level optimization~\cite{colson2007overview} problem:
\begin{align}
\max\nolimits_{\{\Omega|G\}}
F
\left( \{\mathcal{T}_{1}, ..., \mathcal{T}_{L}\}, \Theta^\star \right),
\quad
\textrm{s.t.}
\quad
\begin{cases}
\{\mathcal{T}_{1}, ..., \mathcal{T}_{L}\} = \mathrm{Match}(G, \Omega)
\\
\Theta^\star 
\! = \! \arg\max\nolimits_{\Theta}
F(\{\mathcal{T}_{1}, ..., \mathcal{T}_{L}\},  \Theta)
\end{cases},
\label{eq:minimize}
\end{align}
However, the proposed optimization problem is difficult mainly for two reasons: 1) the search space of subgraph templates is discrete and not differentiable; 2) matching subgraph templates in a large graph is computationally expensive.

Our key intuition to address these challenges is that similar subgraph templates often have slightly different pools of matched instances, which is likely to result in similar model performance. 
Therefore, by gradually exploring the adjacent space of good performing subgraph templates  we can effectively avoid bad candidates. 
This inspires us to design a genetic optimization framework, which can navigate through the discrete search space via the gradual mutations between generations.
Moreover, the similarities between iteratively searched subgraphs can be further leveraged to design efficient subgraph matching algorithm. 
The details are described as follows.

\subsubsection{Genetic Subgraph Template Search}
We define gene population as a set of $B$ \emph{genes}, where each gene is a set of $L$ subgraph templates. The gene population is initiated as the most basic subgraph templates, \emph{i.e.}, edge. 
Then, the gene population is optimized through $K_2$ rounds of genetic operations, which consists of \emph{mutate}, \emph{crossover}, \emph{evaluate} and \emph{select}.
The \emph{mutate} operation allows us to explore slightly more complex subgraph templates, 
\emph{i.e.}, the subgraph templates with one randomly added nodes or edges, which are denoted as \emph{children subgraphs}. 
Besides, the \emph{crossover} operation will randomly exchange some subgraph templates between two genes, which allows us to try different combinations of subgraph templates.
Moreover, the \emph{evaluate} and \emph{select} operations will identify and remove the worst-performing $Z$ genes and reproduce the best-performing genes. Finally, we use the subgraph templates encoded in the best performing \emph{gene} as the input of $F$. Thus, these operations allow us to gradually explore the adjacent search space of promising \emph{genes} and automatically optimize the subgraph templates. The genetic algorithm is presented in Algorithm~\ref{alg:search}. 

\begin{algorithm}[t]
\caption{Genetic Algorithm for Subgraph Template Optimization}
\label{alg:search}
\begin{algorithmic}[1]
\STATE {Input graph $G = (\mathcal{V}, \mathcal{E})$; node feature $\mathcal{X}(v)$; probability of edge mutation, node mutation, crossover $= p_e, p_n, p_c$; size of gene pool $B$; subgraphs per gene $L$; elimination size $Z$; }
\STATE{$genePool$ = InitPool($B$, $L$);}\\
\FOR{$k \in 1, ..., K_2$}
\STATE{$genePool$ = Mutate($genePool$, $p_e$, $p_n$); \qquad \qquad \ \ /*Mutate Subgraph Templates*/}
\STATE{$genePool$ = Crossover($genePool$, $p_c$); \qquad \qquad \quad /*Generate Different Combinations*/}
\FOR{$gene \in genePool$}
\STATE{$\{\mathcal{T}_{1}, ..., \mathcal{T}_{L}\}$ = $\mathrm{Match}$($G$, $gene$) \qquad \qquad \qquad \quad /*Match on Graph.*/}
\STATE{$accuracy$ = $F(\{\mathcal{T}_{1}, ..., \mathcal{T}_{L}\},  \Theta)$ \qquad \qquad \qquad \quad /*Evaluate Performance.*/}
\STATE $metricPool.append(accuracy)$
\ENDFOR
\STATE{$genePool$ = Select($genePool$, $metricPool$, $Z$) \ \ \quad /*Eliminate and Reproduce*/}
\ENDFOR
\RETURN Best performing $gene \in genePool$.
\end{algorithmic}
\end{algorithm}

\subsubsection{Efficient Subgraph Template Matching}
We have the following Proposition~\ref{prop:match} for the matched instances of the mutated \emph{children subgraph}, which can be leveraged to accelerate the $Match$ function in Algorithm~\ref{alg:search}.
The proof is straightforward since the \emph{children subgraphs} are extended from \emph{parent subgraphs} by randomly adding one node or edge. In fact, the adding edge \emph{mutation} effectively acts as a filtering mechanism on matched instances, and the adding node \emph{mutation} effectively grows the matched instances of \emph{parent subgraph}. Therefore, instead of computing the matched instances of \emph{children subgraph} from scratch, the \emph{mutate} operation facilitates us to save significantly amount of computation by reusing and extending the matched instances of the corresponding \emph{parent subgraphs}. As a result, we propose an incremental subgraph matching algorithm to leverage this proposition to accelerate subgraph matching process, which is illustrated in 
Appendix~\ref{app:subgraph_match} in details.

\begin{prop}
\label{prop:match}
Given a graph $G$, a parent subgraph $S_p$ and a children subgraph $S_c$, 
we denote $S_p$ and $S_c$'s match instances set as $\mathcal{M}_p$ and $\mathcal{M}_c$, respectively. Then, we have $m_p \subset m_c$:
$\exists m_p \in \mathcal{M}_p$, for 
$\forall m_c \in \mathcal{M}_c$. 
That is the matched instances of children subgraphs $m_c$ will always contain a matched instance of parent subgraphs $m_p$. Thus, $m_c$ can be efficiently identified by incrementally extending $m_p$.
\end{prop}

\subsection{Theoretical Analysis}
\label{sec:theory}

Here, we aim to answer two questions: 
1) how does the expressiveness of AE-aware aggregator relate to previous works; 
and 2) is our designed AE-aware aggregator expressive enough to capture Ego-AE feature. 
Previous researches mainly investigate the expressiveness of GNN through the scope of graph isomorphism test, 
while it is expressive power on capturing AE feature is largely unknown. 
Specifically, we have the following proposition about the limitations of standard MPNNs.



\begin{prop}
\label{prop:limitation}
There exist graphs that have different Ego-AE sets for a given node, but MPNNs in (\ref{eq:mpnn}) with arbitrary number of layers and hidden units cannot distinguish them.
\end{prop}

We provide a constructive proof in Appendix~\ref{app:limitation}. 
Therefore, this proposition shows MPNNs have fundamental limitations in modeling the structural role of each node's neighbors. 
On the contrary, we have the following theorem about the expressive power of the proposed AE-aware aggregator.
The theoretical proof of this theorem is provided in Appendix~\ref{app:prop_expressive}. It shows our AE-aware aggregator is provably expressive in capturing Ego-AE feature.

\begin{thm}\label{thm:expressive}
For countable feature space $\mathcal{X}$, let $v_a$ and $v_b$ be two nodes with different Ego-AE sets. The AE-aware aggregator in (\ref{eq:gae}) can discriminate two nodes with learned distinct embeddings.
\end{thm}

\begin{remark}
Previous works on GNN's structural feature expressiveness mainly follow the hierarchy of neighborhood isomorphic, \emph{i.e.}, can the GNNs differentiate two nodes that have different isomorphisms in their local neighborhoods. However, graph automorphism is a special isomorphism that maps a graph on to itself~\cite{mckay2014practical}. Therefore, Ego-AE is a stricter structural condition than neighborhood isomorphic. That is two automorphically equivalent nodes are always neighborhood isomporphic, while the converse statement is false even if the local neighborhoods expand to the entire graph~\cite{everett1990ego}. As a result, previous works that aim to differentiate neighborhood isomorphic nodes with various forms of WL tests, \emph{e.g.}, GIN~\cite{xu2018powerful} and k-GNN~\cite{morris2019weisfeiler}, cannot capture Ego-AE feature. Our GRAPE model aims to fill in this gap.
\end{remark}

\section{Experiments}


\noindent
\textbf{Datasets.}
Three types of real-world datasets are used, \emph{i.e.}, academic citation networks, social networks and e-commerce co-purchase network.
\begin{itemize}[leftmargin=0.4cm,topsep=0pt,parsep=0pt,partopsep=0pt]
\item 
\textit{Citation networks}~\cite{sen2008collective}: We consider 2 widely used citation networks, \emph{i.e.} Cora and Citeseer. In these datasets, nodes represent academic papers and (undirected) edges denote the citation links between them. 
Following the setting in previous work~\cite{kipf2016semi}, we use each paper's bag-of-words vector as its node feature and the subject as its label. 

\item
\textit{Social networks}~\cite{traud2012social}: We use 5 real-world social networks which are collected from the Facebook friendships within 5 universities, \emph{i.e.}, Hamilton, Lehigh, Rochester, Johns Hopkins (JHU) and Amherst. The nodes represent students and faculties. Besides, we use one-hot encoding of their gender and major as node feature, and set the labels as their enrollment years.
	
\item
\textit{E-commerce co-purchase networks}~\cite{leskovec2007dynamics}: This dataset was collected by crawling the \emph{music} items in Amazon website. If an item $a$ is frequently co-purchased with $b$, the graph contains a directed edge from $a$ to $b$. We use the average ratings of items as node labels, and we set the node features as the number of reviews and downloads. 
\end{itemize}

\noindent
\textbf{Compared Methods.}
We compare our GRAPE with state-of-the-art GNN models, 
including GCN~\cite{kipf2016semi}, GraphSAGE~\cite{hamilton2017inductive}, GIN~\cite{xu2018powerful}, GAT~\cite{velivckovic2017graph}, 
Geniepath~\cite{liu2019geniepath}, Mixhop~\cite{abu2019mixhop}, Meta-GNN~\cite{sankar2019meta} and DE-GNN~\cite{li2020distance}. 
Specifically, GCN and GraphSAGE are two most popular GNN variants, and GIN is customized to better capture structural property. Besides, GAT and Geniepath use the attention mechanism to learn adaptive neighborhood for each node. Moreover, as more recent baselines, Mixhop, Meta-GNN and DE-GNN learn node embeddings from higher-order structural information. Specifically, Mixhop proposes difference operators on different hops of neighbors;  Meta-GNN leverages predefined subgraphs to identify higher-order neighborhood; DE-GNN encodes the shortest path distance among nodes. To ensure fair comparison, we follow the optimal architectures as described in previous works, and we use the official implementations released by the authors or integrated in Pytorch platform~\cite{paszke2019pytorch}.

Following the common design choices in previous works~\cite{kipf2016semi,hamilton2017inductive,velivckovic2017graph}, we adopt a $2$-layer architecture. 
The hyper-parameter tuning and detailed experiment settings are discussed in Appendix~\ref{app:experiment}. 
Based on the prior knowledge in related areas~\cite{benson2016higher,hanneman2005introduction}, we design five subgraph templates ($S_\star$) for each domain of datasets respectively, which are described in Appendix~\ref{app:subgraph}. We use same subgraph templates for Meta-GNN for fair comparison.

\subsection{Benchmark Comparison}

The classification accuracy of all methods are compared in Table~\ref{tbl:mainresults}.
We observe that GRAPE is the best performing model across all datasets. 
Specifically, the performance gains are most prominent in social datasets. 
The improvements are smaller yet still significant on citation and E-commerce datasets. 
One plausible explanation is the structural features play more important roles in social network analysis. Following~\cite{wang2020gcn,xu2018powerful}, 
to investigate the influence of the node feature on the expressiveness of GNN, we also evaluate the models on datasets that use all-ones dummy node features and randomly initialized node features (see Appendix~\ref{app:dummy} for details). 
We observe that GRAPE achieves consistent performance gains independent of node features, 
which echos the findings in previous studies that stronger topological feature can usually boost GNN's learning performance~\cite{xu2018powerful,you2019position,wang2020gcn}.

\begin{table}[ht]
\caption{Classification accuracy on datasets with original node feature (\%). The best-performing GNNs are in boldface, and the second best ones are underlined.}
\centering
\setlength\tabcolsep{5pt}
\begin{tabular}{p{1.7cm}<{\centering}m{1.15cm}<{\centering}m{1.15cm}<{\centering}m{1.15cm}<{\centering}m{1.15cm}<{\centering}m{1.15cm}<{\centering}m{1.15cm}<{\centering}m{1.15cm}<{\centering}m{1.15cm}<{\centering}}
	\toprule
	                                                                       &                                                 \multicolumn{5}{c} {\textbf{Social}}                                                 &       \multicolumn{2}{c} {\textbf{Citation}}        & \textbf{Ecomm.}          \\
	\cmidrule(lr){2-6} 
	\cmidrule(lr){7-8} 
	\cmidrule(lr){9-9}
	Model & Hamilton                 & Lehigh                   & Rochester                & JHU                      & Amherst                  & Cora                     & Citeseer                 & Amazon                   \\ \midrule
	GCN                                                                    & 19.4$\pm$2.0             & \underline{24.0$\pm$1.2} & 21.1$\pm$1.5             & 20.5$\pm$0.8             & 17.0$\pm$1.4             & \textbf{86.9$\pm$1.7} & \textbf{74.8$\pm$1.0} & 47.4$\pm$1.3             \\
	GraphSAGE                                                              & 20.5$\pm$2.0             & 17.8$\pm$2.2             & 18.5$\pm$1.8             & 17.1$\pm$2.5             & 17.8$\pm$1.6             & 85.6$\pm$0.9             & 70.3$\pm$1.3             & 19.6$\pm$1.0             \\
	GIN                                                                    & \underline{23.7$\pm$3.1} & 19.0$\pm$2.0             & 21.2$\pm$0.9             & 21.5$\pm$3.5             & \underline{26.3$\pm$3.8} & 86.5$\pm$1.2             & 72.6$\pm$1.5             & 48.5$\pm$2.5             \\
	GAT                                                                    & 18.3$\pm$2.1             & 22.7$\pm$0.9             & 20.2$\pm$1.6             & 19.8$\pm$1.4             & 17.7$\pm$2.2             & \textbf{87.1$\pm$2.1}    & \textbf{74.6$\pm$1.7} & 39.4$\pm$0.8             \\
	Geniepath                                                              & 19.1$\pm$1.5             & 22.9$\pm$1.0             & 21.7$\pm$1.1             & 19.8$\pm$1.4             & 17.5$\pm$1.7             & 81.2$\pm$1.5             & 69.8$\pm$1.7             & 57.9$\pm$0.8             \\ \midrule
	Meta-GNN                                                               & \underline{22.9$\pm$1.8} & \underline{24.3$\pm$2.3} & \underline{23.2$\pm$0.5} & \underline{27.3$\pm$2.9} & 23.5$\pm$2.2             & \underline{86.8$\pm$1.1} & \underline{74.4$\pm$1.1} & 56.6$\pm$1.2             \\
	Mixhop                                                                 & 19.1$\pm$0.1             & 23.2$\pm$0.2             & 18.0$\pm$0.0             & 18.3$\pm$0.1             & 17.0$\pm$0.1             & 80.9$\pm$0.7             & 72.9$\pm$0.8             & 57.4$\pm$0.8             \\ \midrule
	DE-GNN                                                                 & 21.7$\pm$2.5             & \underline{24.9$\pm$2.1} & 18.0$\pm$0.0             & 19.8$\pm$1.7             & 18.9$\pm$2.2             & 31.9$\pm$0.0             & 39.1$\pm$2.3             & \underline{58.1$\pm$0.1} \\ \midrule
	GRAPE                                                                  & \textbf{28.1$\pm$2.1}    & \textbf{27.3$\pm$3.8}    & \textbf{25.0$\pm$1.8}    & \textbf{34.6$\pm$1.3}    & \textbf{32.6$\pm$2.2}    & \textbf{87.1$\pm$1.8}    & \textbf{74.6$\pm$1.5}    & \textbf{58.6$\pm$0.4}    \\ \bottomrule
\end{tabular}
\label{tbl:mainresults}
\end{table}

\begin{figure}[t]
	\centering
	\subfigure[Wall-clock time]{\includegraphics[width=0.31\textwidth]{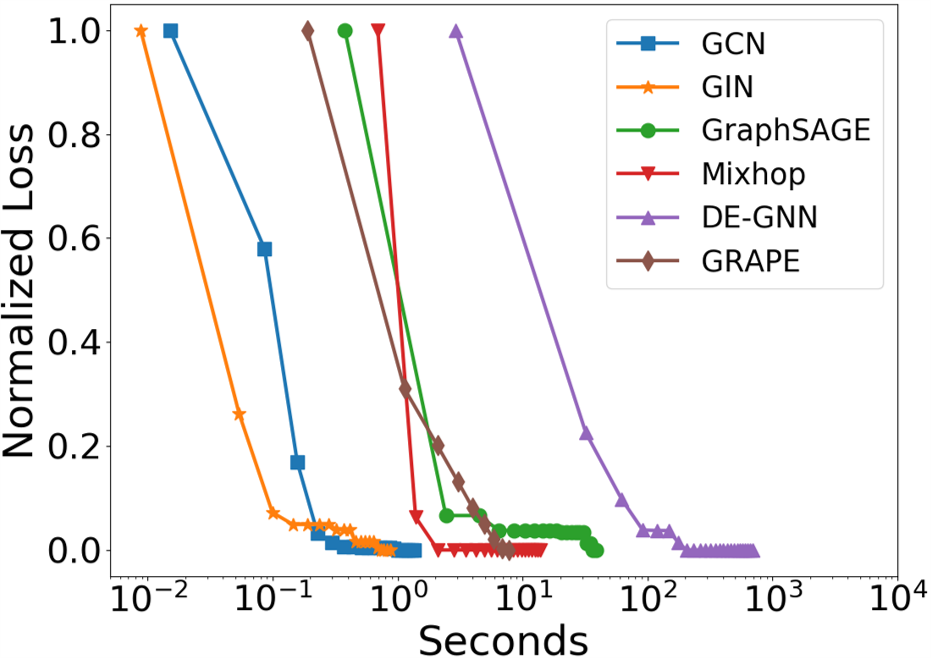}}\quad
	\subfigure[Weights on subgraphs]{\includegraphics[width=0.30\textwidth]{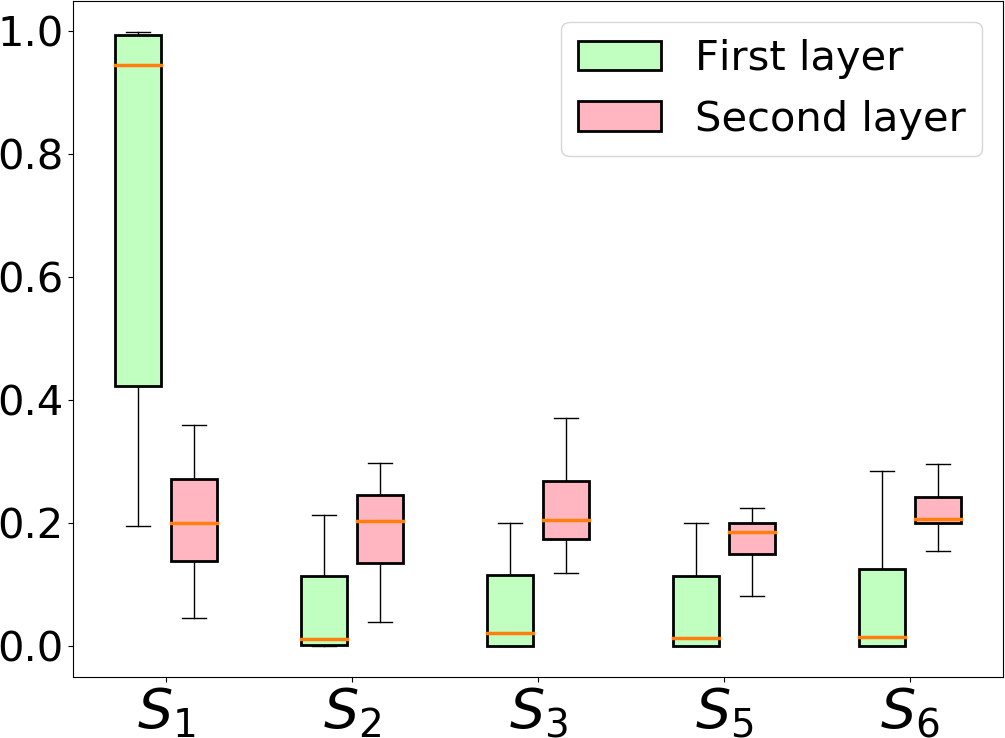}}\quad
	\subfigure[Weights on Ego-AE sets]{\includegraphics[width=0.30\textwidth]{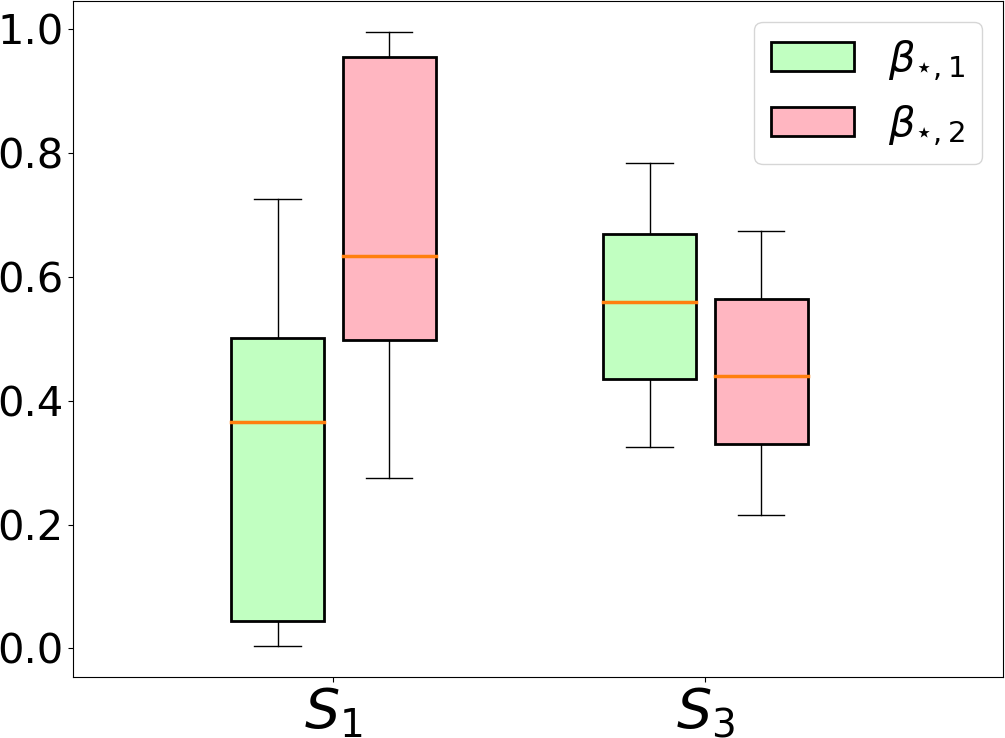}}
	\caption{Illustrating the training time and the attention weights learned by GRAPE on \emph{Lehigh} dataset.}
	\label{fig:attn}
\end{figure}

We show the computation cost of GRAPE and exemplar GNNs in terms of wall-clock training time in Figure~\ref{fig:attn}~(a). 
Note that both Geniepath and GAT leverage attention mechanism, which can only be trained on CPU (the training with GPU causes the out-of-memory error). Meta-GNN is coupled with complex graph sampling process and takes much more time to train. Thus, these methods are not plotted. 
We observe GRAPE takes comparable time to train on the example dataset as the classic GNN variants of Mixhop and GraphSAGE, which demonstrates the efficiency of our model. The theoretical time complexity analysis is provided in Appendix~\ref{app:time}.

Besides, we also analyze the learned squeeze-and-excitation weights on each subgraph template and Ego-AE set.
From Figure~\ref{fig:attn}~(b), 
we observe the weight is significantly skewed to edge template $S_1$ in first layer, 
while it distributes more evenly on more complex templates in the second layer. One plausible reason is that the model tend to keep the neighborhood large in the first hop to collect more feature. As a result, it assigns higher weights to the simplest template, \emph{i.e.}, $S_1$ edge, in the first layer, while it fuses more diverse structural feature in the second layer by assigning more even weights to various subgraph templates. 
Figure~\ref{fig:attn} (c) shows the AE-aware aggregator can effectively distinguish the neighboring nodes with different structural roles. For example, in \emph{Lehigh} dataset, 
GRAPE assigns lower weights to the Ego-AE set of ego node $\{u_1\}$ and higher weights to the AE set of neighbors $\{u_2\}$ on edge template $S_1$, while the weights distribute more evenly between $\{u_1\}$ and $\{u_2, u_3\}$ on triangle template $S_3$. It suggests the connected neighborhood nodes are more important than ego node on edge template $S_1$, while they have similar importance on triangle template $S_3$.

\subsection{Case Study on AE-aware Aggregators}

\begin{figure}[ht]
\centering
\subfigure[2-hop neighborhood]
{\includegraphics[width=0.25\columnwidth]{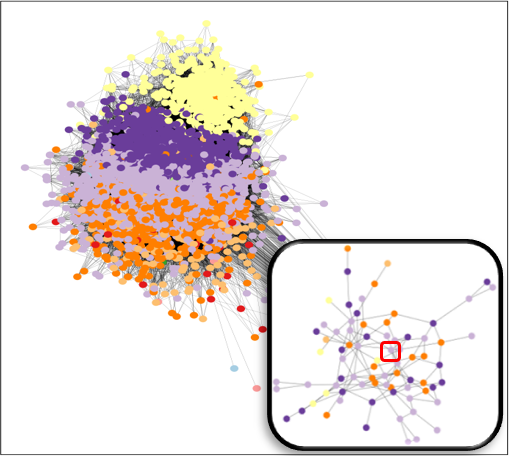}}
\qquad
\subfigure[Matched with $S_3$]
{\includegraphics[width=0.25\columnwidth]{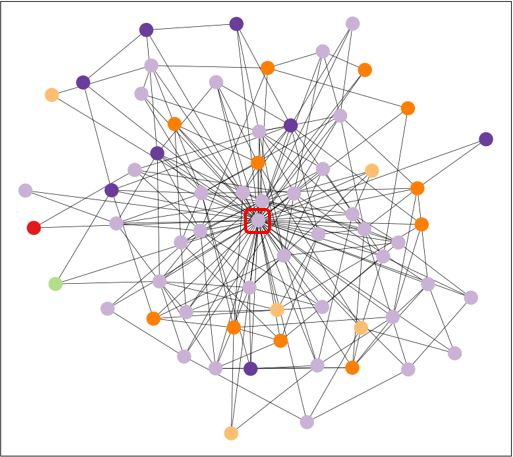}}
\qquad
\subfigure[Matched with $S_5$]
{\includegraphics[width=0.25\columnwidth]{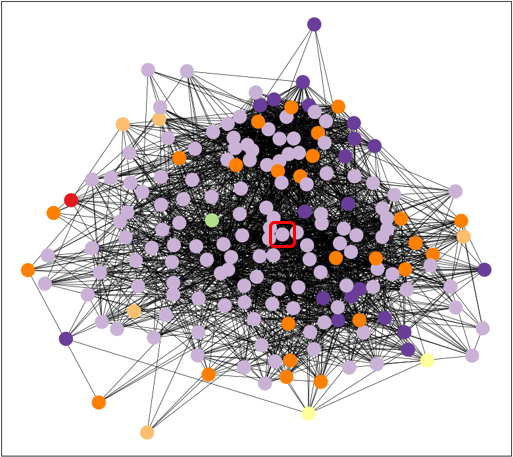}}
\caption{Case study of a node $v_c$ in \emph{Lehigh} dataset. The ego node $v_c$ is positioned in the center and the inset of (a) shows 100 samples. Node colors represent the ground-truth label.}
\label{fig:badcase}
\end{figure}

We conduct a case study to better understand how the AE-aware aggregators contribute to GRAPE. 
Specifically, we first randomly select a node $v_c$ in \emph{Lehigh} dataset that is wrongly classified by all GNN variants except GRAPE, 
and then analyze its 2-hop neighborhood $\mathcal{N}^2(v_c)$ and the neighboring nodes that match triangle template $S_3$ and 4-clique template $S_5$ in Figure~\ref{fig:badcase} and Table~\ref{tbl:bad}. 
We observe that there are 3,600 nodes in $\mathcal{N}^2(v_c)$ and they distribute evenly on multiple labels, which not only might confuse MPNNs but also tend to cause over-smoothness problem~\cite{li2018deeper}. 
On the other hand, both templates $S_3$ and $S_5$ significantly reduce the neighborhood size, and the percentage of neighbors that have same labels as $v_c$ increases from 19.7\% in $\mathcal{N}^2(v_c)$ (ranked $3^{rd}$) to 45.3\%$\sim$46.0\% in $\mathcal{A}_{3,2}(v_c)$ and $\mathcal{A}_{5,2}(v_c)$ (ranked $1^{st}$). 
It shows the AE-aware aggregator can successfully capture the ``social homophily'' effect, where the nodes in tightly connected communities, \emph{e.g.}, triangle and 4-clique structure, tend to have similar property~\cite{mcpherson2001birds}.

\begin{table}[ht]
\caption{Number of nodes and most frequent labels in $v_c$'s 2-hop neighborhood and Ego-AE sets.}
\centering
\setlength\tabcolsep{4pt}
\begin{tabular}{c|c|c}
\toprule
& \# nodes &The percentage of top 5 labels ($v_c$'s label is 11)\\
\hline
2-hop neigh. $\mathcal{N}^2(v_c)$ & 3,600& 13 (23.2\%), 12 (21.8\%), \textbf{11 (19.7\%)}, 10 (17.3\%), 14 (12.9\%)\\
Ego-AE set $\mathcal{A}_{3,2} (v_c)$ & 46& \textbf{11 (45.3\%)}, 10 (17.2\%), 12 (14.1\%), 13 (12.5\%), 9 (7.8\%)\\
Ego-AE set $\mathcal{A}_{5,2} (v_c)$ & 222& \textbf{11 (46.0\%)}, 12 (24.9\%), 10 (13.8\%), 13 (11.0\%), 9 (2.8\%)\\ 
\bottomrule
\end{tabular}
\label{tbl:bad}
\end{table}

\subsection{Subgraph Template Search}

\begin{figure}[ht]
\centering
\begin{minipage}{0.29\textwidth}
Figure~\ref{fig:search} shows a case study of efficiency and searched subgraph templates of Algorithm~\ref{alg:search}. 
We observe that Algorithm~\ref{alg:search} can generate similar prediction accuracy compared to the hand-craft subgraph templates in reasonable search time.
Moreover, these subgraph templates are data-dependent. 
\end{minipage}
\;\;
\begin{minipage}{0.68\textwidth}
\includegraphics[width=0.51\textwidth]{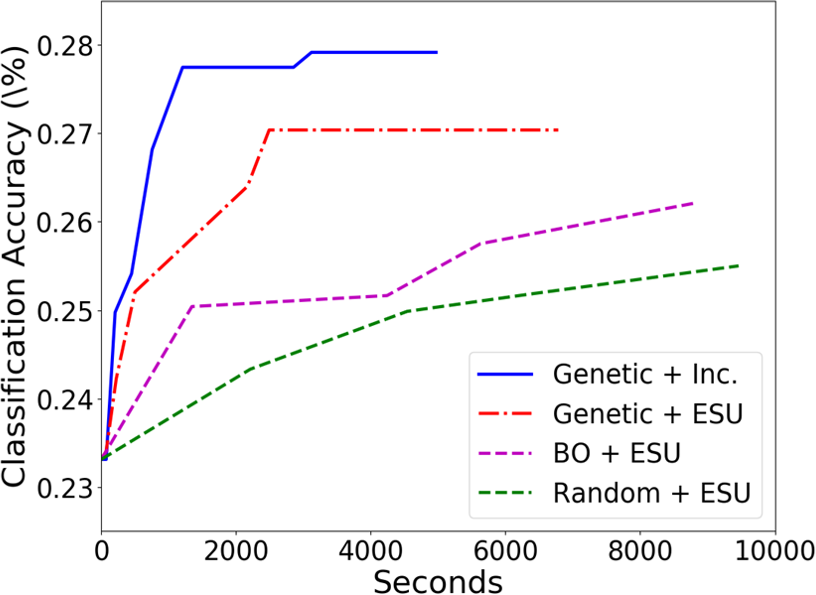}
\includegraphics[width=0.48\textwidth]{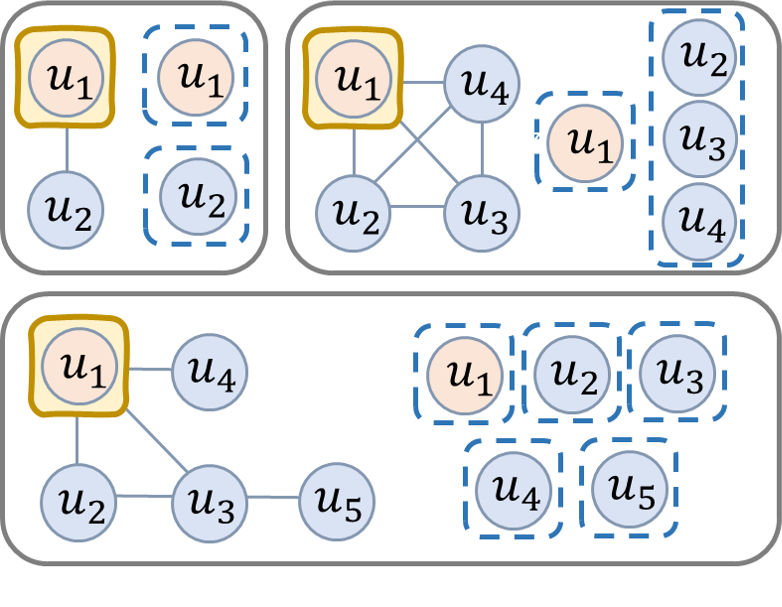}
\captionof{figure}{Effectiveness of the proposed genetic algorithm on the \emph{Lehigh} Dataset.
	Left: search efficiency;
	right: subgraph templates.}
\label{fig:search}
\end{minipage}
\end{figure}

We evaluate the genetic subgraph template search algorithm on the social dataset. 
Specifically, we initialize GRAPE with three simplest subgraph template, \emph{i.e.}, edges. We compare our algorithm with two baseline methods: 1) Genetic + ESU, which replaces the incremental subgraph matching algorithm in our genetic framework with a widely adopted baseline method, \emph{i.e.}, \emph{ESU}~\cite{wernicke2006fanmod}; 2) Random + ESU: search randomly initialized subgraph templates with \emph{ESU}. Besides, we also add a Bayesian Optimization (BO) + ESU baseline, which uses the the classic Bayesian optimization to search for optimal subgraph templates~\cite{yao2018taking}. We use the BO model implemented in Hyperopt framework~\cite{bergstra2013hyperopt}. The optimization results are shown in Figure~\ref{fig:search} and Table~\ref{tbl:search}. 
Specifically, Table~\ref{tbl:search} shows the proposed Genetic method (Genetic + INC.) outperform all three baselines across all the datasets. Our genetic framework generates 3.0$\sim$8.0\% performance gain over the initialized simple edge template within 3000 seconds, which is 0.6$\sim$2.8\% higher compared with the best performing baseline. Moreover, the optimized performance is comparable with the hand-crafted subgraph templates in Table~\ref{tbl:mainresults}. 

\begin{table}[ht]
\caption{Classification accuracy (\%) and time composition after 3000 seconds genetic optimization.}
\centering
\setlength\tabcolsep{3pt}
\begin{tabular}{p{2cm}<{\centering}|c|c c c c c}
	\toprule
	\multicolumn{2}{c|}{\ }                                          & Hamilton & Lehigh & Rochester & JHU    & Amherst \\ \midrule
	\multirow{4}*{\shortstack{\textbf{Classification} \\ \textbf{Accuracy (\%)}}}          & Edge Template (Init.)          &   25.8   & 23.3   & 22.6      & 26.0   & 28.3    \\
	& Random+ESU       &   29.2   & 24.5   & 24.8      & 26.0   & 28.3    \\
	&BO+ESU       &   30.1   & 25.0   & 24.6      & 26.2   & 29.5    \\                                                                & Genetic+ESU      &   31.0   & 27.0   & 25.0      & 27.5   & 33.0    \\
	& Genetic+INC.     &   33.8   & 27.9   & 25.6      & 30.3   & 34.5    \\ \midrule
	\multirow{3}*{\shortstack{\textbf{Time} \\ \textbf{Composition}\\ \textbf{(Seconds)}}} & ESU Matching     &  648.7   & 889.4  & 2592.0    & 1389.2 & 1357.1  \\
	& INC. Matching    &   24.0   & 55.3   & 134.3     & 129.9  & 124.6   \\
	& Model Evaluation &   20.9   & 33.7   & 45.0      & 31.5   & 18.1    \\ \bottomrule
\end{tabular}
\label{tbl:search}
\end{table}

We also evaluate the performance of various hand-craft subgraph templates in Appendix~\ref{app:result_subgraph}. We can see that the classification accuracy on \emph{Lehigh} dataset ranges from 22.6\% to 26.1\%, which show the importance of subgraph templates and the need for automatic subgraph template search.

\section{Conclusion}

In this paper, we design a theoretical framework to examine GNN's expressiveness through the lens of Ego-AE. We prove MPNNs have fundamental limitations in capturing this important structural property. 
Moreover, we propose a provably expressive GNN model, \emph{i.e.}, GRAPE, which effectively extend GNN's capability in modeling structural roles. 
We also design a genetic subgraph template search algorithm to automatically optimize the model architecture. Experiments on real-world datasets show consistent performance gain of the proposed methods.
One potential limitation of our model is the scalability to large-scale real-world graphs (see complexity analysis in Appendix~\ref{app:time}). However, since our model is defined on localized Ego-AE, sampling technique similar to GraphSage~\cite{hamilton2017inductive} can be designed to ensure the feasibility of computation overhead, 
which we will leave as a future work. 
Besides, our paper proposes a general GNN framework and it could be customized for specific application domains, \emph{e.g.}, molecular property prediction~\cite{wang2021property}, 
to achieve additional performance gain.

\clearpage


\bibliographystyle{plain}
\bibliography{neurips}

\begin{thebibliography}{10}

\bibitem{abu2019mixhop}
Sami Abu-El-Haija, Bryan Perozzi, Amol Kapoor, Nazanin Alipourfard, Kristina
  Lerman, Hrayr Harutyunyan, Greg Ver~Steeg, and Aram Galstyan.
\newblock Mixhop: Higher-order graph convolutional architectures via sparsified
  neighborhood mixing.
\newblock In {\em ICML}, pages 21--29. PMLR, 2019.

\bibitem{ahmed2020role}
Nesreen Ahmed, Ryan~Anthony Rossi, John Lee, Theodore Willke, Rong Zhou,
  Xiangnan Kong, and Hoda Eldardiry.
\newblock Role-based graph embeddings.
\newblock {\em IEEE TKDE}, 2020.

\bibitem{benson2016higher}
Austin~R Benson, David~F Gleich, and Jure Leskovec.
\newblock Higher-order organization of complex networks.
\newblock {\em Science}, 353(6295):163--166, 2016.

\bibitem{bergstra2013hyperopt}
James Bergstra, Dan Yamins, David~D Cox, et~al.
\newblock Hyperopt: A python library for optimizing the hyperparameters of
  machine learning algorithms.
\newblock In {\em Python in Science Conference}, volume~13, page~20. Citeseer,
  2013.

\bibitem{bouritsas2020improving}
Giorgos Bouritsas, Fabrizio Frasca, Stefanos Zafeiriou, and Michael~M
  Bronstein.
\newblock Improving graph neural network expressivity via subgraph isomorphism
  counting.
\newblock {\em arXiv preprint arXiv:2006.09252}, 2020.

\bibitem{colbourn1981linear}
Charles~J Colbourn and Kellogg~S Booth.
\newblock Linear time automorphism algorithms for trees, interval graphs, and
  planar graphs.
\newblock {\em SIAM Journal on Computing}, 10(1):203--225, 1981.

\bibitem{colson2007overview}
Beno{\^\i}t Colson, Patrice Marcotte, and Gilles Savard.
\newblock An overview of bilevel optimization.
\newblock {\em Annals of operations research}, 153(1):235--256, 2007.

\bibitem{cordella2004sub}
Luigi~P Cordella, Pasquale Foggia, Carlo Sansone, and Mario Vento.
\newblock A (sub) graph isomorphism algorithm for matching large graphs.
\newblock {\em IEEE TPAMI}, 26(10):1367--1372, 2004.

\bibitem{corso2020principal}
Gabriele Corso, Luca Cavalleri, Dominique Beaini, Pietro Li{\`o}, and Petar
  Veli{\v{c}}kovi{\'c}.
\newblock Principal neighbourhood aggregation for graph nets.
\newblock {\em arXiv preprint arXiv:2004.05718}, 2020.

\bibitem{darwin1909origin}
Charles Darwin.
\newblock {\em The origin of species}.
\newblock PF Collier \& son New York, 1909.

\bibitem{dehmamy2019understanding}
Nima Dehmamy, Albert-L{\'a}szl{\'o} Barab{\'a}si, and Rose Yu.
\newblock Understanding the representation power of graph neural networks in
  learning graph topology.
\newblock In {\em NeurIPS}, pages 15387--15397, 2019.

\bibitem{donnat2018learning}
Claire Donnat, Marinka Zitnik, David Hallac, and Jure Leskovec.
\newblock Learning structural node embeddings via diffusion wavelets.
\newblock In {\em SIGKDD}, pages 1320--1329, 2018.

\bibitem{eiben2003introduction}
Agoston~E Eiben, James~E Smith, et~al.
\newblock {\em Introduction to evolutionary computing}, volume~53.
\newblock Springer, 2003.

\bibitem{everett1985role}
Martin~G Everett.
\newblock Role similarity and complexity in social networks.
\newblock {\em Social Networks}, 7(4):353--359, 1985.

\bibitem{everett1990ego}
Martin~G Everett, John~P Boyd, and Stephen~P Borgatti.
\newblock Ego-centered and local roles: A graph theoretic approach.
\newblock {\em Journal of Mathematical Sociology}, 15(3-4):163--172, 1990.

\bibitem{friedkin1997social}
Noah~E Friedkin and Eugene~C Johnsen.
\newblock Social positions in influence networks.
\newblock {\em Social Networks}, 19(3):209--222, 1997.

\bibitem{garg2020generalization}
Vikas~K Garg, Stefanie Jegelka, and Tommi Jaakkola.
\newblock Generalization and representational limits of graph neural networks.
\newblock {\em arXiv preprint arXiv:2002.06157}, 2020.

\bibitem{gilmer2017neural}
Justin Gilmer, Samuel~S Schoenholz, Patrick~F Riley, Oriol Vinyals, and
  George~E Dahl.
\newblock Neural message passing for quantum chemistry.
\newblock In {\em ICML}, pages 1263--1272. JMLR. org, 2017.

\bibitem{hamilton2017inductive}
Will Hamilton, Zhitao Ying, and Jure Leskovec.
\newblock Inductive representation learning on large graphs.
\newblock In {\em NIPS}, pages 1024--1034, 2017.

\bibitem{hanneman2005introduction}
Robert~A Hanneman and Mark Riddle.
\newblock Introduction to social network methods, 2005.

\bibitem{hornik1989multilayer}
Kurt Hornik, Maxwell Stinchcombe, Halbert White, et~al.
\newblock Multilayer feedforward networks are universal approximators.
\newblock {\em NN}, 2(5):359--366, 1989.

\bibitem{hu2018squeeze}
Jie Hu, Li~Shen, and Gang Sun.
\newblock Squeeze-and-excitation networks.
\newblock In {\em CVPR}, pages 7132--7141, 2018.

\bibitem{huang2014mining}
Hong Huang, Jie Tang, Sen Wu, Lu~Liu, and Xiaoming Fu.
\newblock Mining triadic closure patterns in social networks.
\newblock In {\em WWW}, pages 499--504, 2014.

\bibitem{kipf2016semi}
Thomas~N Kipf and Max Welling.
\newblock Semi-supervised classification with graph convolutional networks.
\newblock In {\em ICLR}, 2016.

\bibitem{klicpera2020directional}
Johannes Klicpera, Janek Gro{\ss}, and Stephan G{\"u}nnemann.
\newblock Directional message passing for molecular graphs.
\newblock {\em arXiv preprint arXiv:2003.03123}, 2020.

\bibitem{leskovec2007dynamics}
Jure Leskovec, Lada~A Adamic, and Bernardo~A Huberman.
\newblock The dynamics of viral marketing.
\newblock {\em ACM TWEB}, 1(1):5--es, 2007.

\bibitem{li2020distance}
Pan Li, Yanbang Wang, Hongwei Wang, and Jure Leskovec.
\newblock Distance encoding: Design provably more powerful neural networks for
  graph representation learning.
\newblock In {\em NeurIPS}, volume~33, 2020.

\bibitem{li2018deeper}
Qimai Li, Zhichao Han, and Xiao-Ming Wu.
\newblock Deeper insights into graph convolutional networks for semi-supervised
  learning.
\newblock In {\em AAAI}, 2018.

\bibitem{lin2021learning}
Zhi-Hao Lin, Sheng~Yu Huang, and Yu-Chiang~Frank Wang.
\newblock Learning of 3d graph convolution networks for point cloud analysis.
\newblock {\em IEEE TPAMI}, 2021.

\bibitem{liu2021graph}
Yixin Liu, Shirui Pan, Ming Jin, Chuan Zhou, Feng Xia, and Philip~S Yu.
\newblock Graph self-supervised learning: A survey.
\newblock {\em arXiv preprint arXiv:2103.00111}, 2021.

\bibitem{liu2019geniepath}
Ziqi Liu, Chaochao Chen, Longfei Li, Jun Zhou, Xiaolong Li, Le~Song, and Yuan
  Qi.
\newblock Geniepath: Graph neural networks with adaptive receptive paths.
\newblock In {\em AAAI}, volume~33, pages 4424--4431, 2019.

\bibitem{lorenzo2017particle}
Pablo~Ribalta Lorenzo, Jakub Nalepa, Michal Kawulok, Luciano~Sanchez Ramos, and
  Jos{\'e}~Ranilla Pastor.
\newblock Particle swarm optimization for hyper-parameter selection in deep
  neural networks.
\newblock In {\em GECCO}, pages 481--488, 2017.

\bibitem{lorrain1971structural}
Francois Lorrain and Harrison~C White.
\newblock Structural equivalence of individuals in social networks.
\newblock {\em Journal of Mathematical Sociology}, 1(1):49--80, 1971.

\bibitem{luczkovich2003defining}
Joseph~J Luczkovich, Stephen~P Borgatti, Jeffrey~C Johnson, and Martin~G
  Everett.
\newblock Defining and measuring trophic role similarity in food webs using
  regular equivalence.
\newblock {\em Journal of Theoretical Biology}, 220(3):303--321, 2003.

\bibitem{mckay2014practical}
Brendan~D McKay and Adolfo Piperno.
\newblock Practical graph isomorphism, ii.
\newblock {\em Journal of Symbolic Computation}, 60:94--112, 2014.

\bibitem{mcpherson2001birds}
Miller McPherson, Lynn Smith-Lovin, and James~M Cook.
\newblock Birds of a feather: Homophily in social networks.
\newblock {\em Annual review of sociology}, 27(1):415--444, 2001.

\bibitem{mizruchi1993cohesion}
Mark~S Mizruchi.
\newblock Cohesion, equivalence, and similarity of behavior: A theoretical and
  empirical assessment.
\newblock {\em Social Networks}, 15(3):275--307, 1993.

\bibitem{morris2019weisfeiler}
Christopher Morris, Martin Ritzert, Matthias Fey, William~L Hamilton, Jan~Eric
  Lenssen, Gaurav Rattan, and Martin Grohe.
\newblock Weisfeiler and leman go neural: Higher-order graph neural networks.
\newblock In {\em AAAI}, volume~33, pages 4602--4609, 2019.

\bibitem{paszke2019pytorch}
Adam Paszke, Sam Gross, Francisco Massa, Adam Lerer, James Bradbury, Gregory
  Chanan, Trevor Killeen, Zeming Lin, Natalia Gimelshein, Luca Antiga, et~al.
\newblock Pytorch: An imperative style, high-performance deep learning library.
\newblock In {\em NeurIPS}, pages 8024--8035, 2019.

\bibitem{ribeiro2017struc2vec}
Leonardo~FR Ribeiro, Pedro~HP Saverese, and Daniel~R Figueiredo.
\newblock struc2vec: Learning node representations from structural identity.
\newblock In {\em SIGKDD}, pages 385--394, 2017.

\bibitem{sankar2019meta}
Aravind Sankar, Xinyang Zhang, and Kevin Chen-Chuan Chang.
\newblock {Meta-GNN}: Metagraph neural network for semi-supervised learning in
  attributed heterogeneous information networks.
\newblock In {\em ASONAM}, pages 137--144, 2019.

\bibitem{sato2019approximation}
Ryoma Sato, Makoto Yamada, and Hisashi Kashima.
\newblock Approximation ratios of graph neural networks for combinatorial
  problems.
\newblock In {\em NeurIPS}, pages 4083--4092, 2019.

\bibitem{sato2021random}
Ryoma Sato, Makoto Yamada, and Hisashi Kashima.
\newblock Random features strengthen graph neural networks.
\newblock In {\em Proceedings of the 2021 SIAM International Conference on Data
  Mining (SDM)}, pages 333--341. SIAM, 2021.

\bibitem{sen2008collective}
Prithviraj Sen, Galileo Namata, Mustafa Bilgic, Lise Getoor, Brian Galligher,
  and Tina Eliassi-Rad.
\newblock Collective classification in network data.
\newblock {\em AI Magazine}, 29(3):93--93, 2008.

\bibitem{sparrow1993linear}
Malcolm~K Sparrow.
\newblock A linear algorithm for computing automorphic equivalence classes: the
  numerical signatures approach.
\newblock {\em Social Networks}, 15(2):151--170, 1993.

\bibitem{toran2004hardness}
Jacobo Tor{\'a}n.
\newblock On the hardness of graph isomorphism.
\newblock {\em SIAM Journal on Computing}, 33(5):1093--1108, 2004.

\bibitem{traud2012social}
Amanda~L Traud, Peter~J Mucha, and Mason~A Porter.
\newblock Social structure of facebook networks.
\newblock {\em Physica A: Statistical Mechanics and its Applications},
  391(16):4165--4180, 2012.

\bibitem{velivckovic2017graph}
Petar Veli{\v{c}}kovi{\'c}, Guillem Cucurull, Arantxa Casanova, Adriana Romero,
  Pietro Lio, and Yoshua Bengio.
\newblock Graph attention networks.
\newblock Technical report, arXiv preprint arXiv:1710.10903, 2017.

\bibitem{wang2020gcn}
Xiao Wang, Meiqi Zhu, Deyu Bo, Peng Cui, Chuan Shi, and Jian Pei.
\newblock Am-gcn: Adaptive multi-channel graph convolutional networks.
\newblock In {\em SIGKDD}, pages 1243--1253, 2020.

\bibitem{wang2021property}
Yaqing Wang, Quanming Abuduweili, Abulikemu~Yao, and Dejing Dou.
\newblock Property-aware relation networks for few-shot molecular property
  prediction.
\newblock In {\em NeurIPS}, 2021.

\bibitem{wernicke2006fanmod}
Sebastian Wernicke and Florian Rasche.
\newblock Fanmod: a tool for fast network motif detection.
\newblock {\em Bioinformatics}, 22(9):1152--1153, 2006.

\bibitem{wu2020comprehensive}
Zonghan Wu, Shirui Pan, Fengwen Chen, Guodong Long, Chengqi Zhang, and S~Yu
  Philip.
\newblock A comprehensive survey on graph neural networks.
\newblock {\em IEEE TNNLS}, 2020.

\bibitem{xie2017genetic}
Lingxi Xie and Alan Yuille.
\newblock Genetic cnn.
\newblock In {\em Proceedings of the IEEE international conference on computer
  vision}, pages 1379--1388, 2017.

\bibitem{xie2021self}
Yaochen Xie, Zhao Xu, Jingtun Zhang, Zhengyang Wang, and Shuiwang Ji.
\newblock Self-supervised learning of graph neural networks: A unified review.
\newblock {\em arXiv preprint arXiv:2102.10757}, 2021.

\bibitem{xu2018powerful}
Keyulu Xu, Weihua Hu, Jure Leskovec, and Stefanie Jegelka.
\newblock How powerful are graph neural networks?
\newblock {\em arXiv preprint arXiv:1810.00826}, 2018.

\bibitem{yao2018taking}
Quanming Yao, Mengshuo Wang, Yuqiang Chen, Wenyuan Dai, Yu-Feng Li, Wei-Wei Tu,
  Qiang Yang, and Yang Yu.
\newblock Taking human out of learning applications: A survey on automated
  machine learning.
\newblock {\em arXiv preprint arXiv:1810.13306}, 2018.

\bibitem{yao1999evolving}
Xin Yao.
\newblock Evolving artificial neural networks.
\newblock {\em Proceedings of the IEEE}, 87(9):1423--1447, 1999.

\bibitem{ying2019gnn}
Rex Ying, Dylan Bourgeois, Jiaxuan You, Marinka Zitnik, and Jure Leskovec.
\newblock Gnn explainer: A tool for post-hoc explanation of graph neural
  networks.
\newblock {\em arXiv preprint arXiv:1903.03894}, 2019.

\bibitem{you2019position}
Jiaxuan You, Rex Ying, and Jure Leskovec.
\newblock Position-aware graph neural networks.
\newblock In {\em ICML}, pages 7134--7143, 2019.

\bibitem{zaheer2017deep}
Manzil Zaheer, Satwik Kottur, Siamak Ravanbakhsh, Barnabas Poczos, Russ~R
  Salakhutdinov, and Alexander~J Smola.
\newblock Deep sets.
\newblock In {\em NIPS}, pages 3391--3401, 2017.

\end{thebibliography}

\cleardoublepage
\appendix



\setcounter{figure}{0}
\renewcommand{\thefigure}{A\arabic{figure}}
\setcounter{table}{0}
\renewcommand{\thetable}{A\arabic{table}}
\setcounter{algorithm}{0}
\renewcommand{\thealgorithm}{A\arabic{algorithm}}

\section{Comparison with Existing GNN Variants}
\label{app:relate}

As described in Section~\ref{sec:rel:MPNN}, 
there are two directions to augment the expressive power of MPNNs: augmenting node features and designing novel architectures. However, we show in Proposition~\ref{prop:limitation} that the important graph property of Ego-AE cannot be captured by the classical MPNN framework, which has not been explored by existing GNN variants. Specifically, previous efforts in feature augmented GNN variants aimed to improve the power by incorporating various additional feature, \emph{e.g.}, graph position~\cite{you2019position,li2020distance}, spatial orientation of edges~\cite{klicpera2020directional} and port numbering~\cite{sato2019approximation}. However, these additional features are often difficult to generalized~\cite{bouritsas2020improving} and they do not investigate the Ego-AE property. In terms of novel architecture, \emph{GIN}~\cite{xu2018powerful} and \emph{k-GNN}~\cite{morris2019weisfeiler} were proposed to optimize the expressiveness in graph isomorphism test, which followed the hierarchy of \emph{1-WL} and \emph{k-WL} framework, respectively. Recent work showed that combining multiple aggregator functions can also improve the expressive power~\cite{corso2020principal}. However, these works mainly investigated the expressiveness of GNNs with graph isomorphism test, which is proven to be an easier task than Ego-AE in previous work~\cite{toran2004hardness}.

Our work follows the later branch of research. Specifically, we propose a novel GNN model, \emph{i.e.}, GRAPE, which can theoretically capture the structural roles defined by Ego-AE. Moreover, we design a genetic algorithm and a compatible incremental subgraph matching algorithm to efficiently search the architecture of GRAPE, which allows it to automatically focus on the most relevant Ego-AE feature in given datasets. To conclude, the proposed GRAPE fundamentally extends GNN's capability in modeling automorphic equivalences and reduces the barrier of generalizing to different datasets. 

\begin{figure}[ht]
\centering
\subfigure[Example graph $G_1$.]{\includegraphics[width = 0.20\textheight]{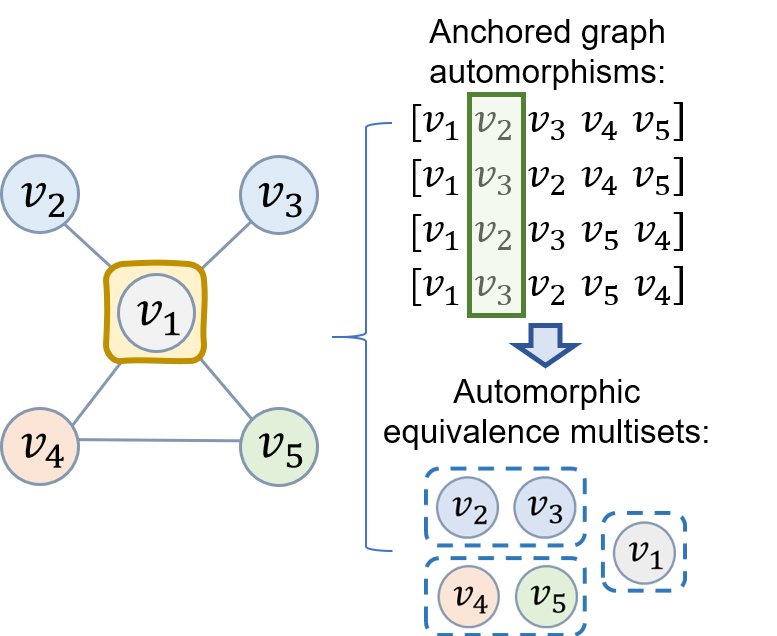}}
\subfigure[Example graph $G_2$.]{\includegraphics[width = 0.20\textheight]{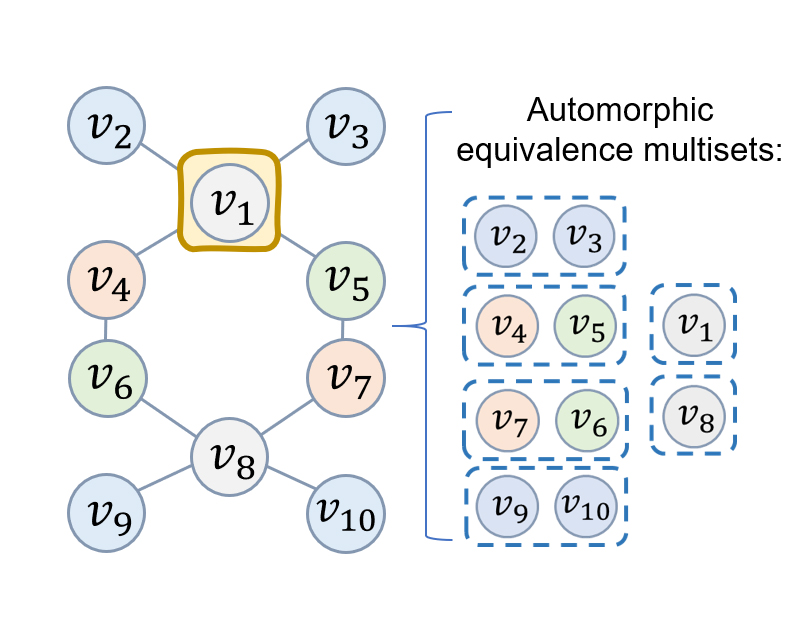}}
\subfigure[Computation graph of MPNN.]{\includegraphics[width = 0.20\textheight]{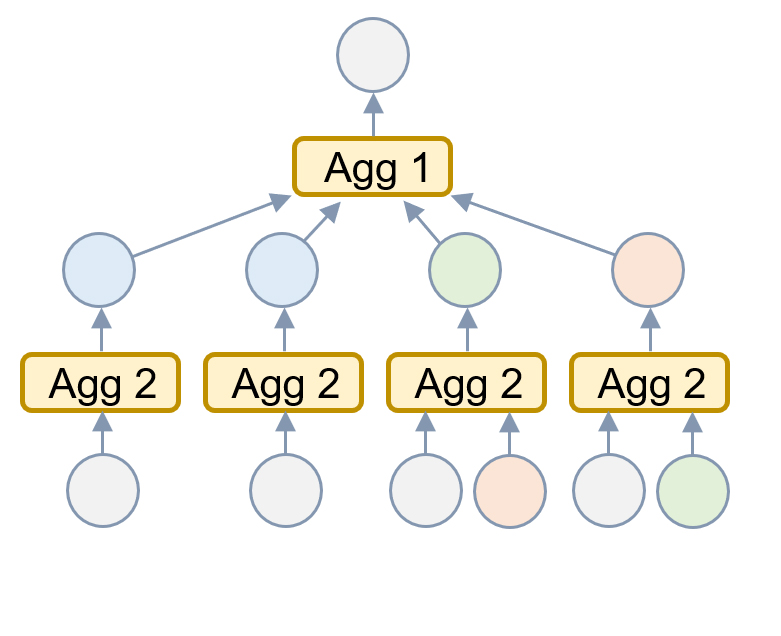}}
\caption{An illustration of the Ego-AE sets in example graphs and the limitations of current GNNs, where the nodes with same colors have identical features.}
\label{fig:automorphism}
\end{figure}

\section{Proofs}

\subsection{Proposition~\ref{prop:limitation}}
\label{app:limitation}

\begin{proof}
We provide a constructive proof for Proposition~\ref{prop:limitation} in Figure~\ref{fig:automorphism}. 
We can observe that example graphs $G_1$ and $G_2$ have different Ego-AE sets for the corresponding ego node $v_1$, 
but their computation graphs of 2-layer GNN are exactly the same, which are shown in Figure~\ref{fig:automorphism} (c). 
In fact, since each node in graph $G_2$ can be mapped to a node in graph $G_1$ with identical 1-hop neighborhood, 
GNNs with arbitrary layers cannot discriminate these two nodes. 
Therefore, it constitutes a constructive proof for Proposition~\ref{prop:limitation}.

\end{proof}

\subsection{Theorem~\ref{thm:expressive}}
\label{app:prop_expressive}

\emph{Proof.} Suppose node $v_a$ and $v_b$ have different Ego-AE sets $\mathcal{T}^a = \{\mathcal{A}^a_1, ...\mathcal{A}^a_j,...\}$ and $\mathcal{T}^b = \{\mathcal{A}^b_1, ..., \mathcal{A}^b_j, ...\}$. Without loss of generality, we assume $\mathcal{A}^a_j$ and $\mathcal{A}^b_j$ are the two sets of nodes that are different. The recently developed ``deep set'' theory provides a framework for injective functions on set data~\cite{zaheer2017deep}, which is then extended to set scenario showing sum operator is an injective function on set~\cite{xu2018powerful}. Therefore, $\textsf{SUM}(\cdot)$ will map them to distinct embeddings $\bm{y}^a_i$ and $\bm{y}^b_j$ since it is an injective function.

Since node feature $\mathcal{X}$ is countable, the embedding of Ego-AE sets $\bm{y}$ is also countable. Therefore, it can be mapped to natural numbers with some function $Z:\mathcal{Y}\rightarrow \mathbb{N}$. Each node has a set of embeddings corresponding to its Ego-AE sets $Y = \{\textsf{SUM}(\{\mathcal{X}(v)| v \in \mathcal{A}_j\})\ |\ \mathcal{A}_j \in \mathcal{T}\}, Y =\{\bm{y}_j\} \subset \mathcal{Y}$, where the cardinality of $Y$ is defined by the number of Ego-AE sets $M$ for the given subgraph template. We can construct a function $f(\bm{y}) = M^{-Z(\bm{y})}$ so that $\sum_{j\in[1,M]} \beta_j f(\bm{y}_j), \bm{y}_j\in Y$ is unique for each set of embeddings, \emph{i.e.}, $\sum_{j\in[1,M]} \beta_j f(\cdot)$ is an injective function on $Y$~\cite{zaheer2017deep}. 

Therefore, for any injective function $g(\cdot)$, the $g(\sum_{j\in[1,M]} \beta_j f(\textsf{SUM}( \{\mathcal{X}(v) | v \in \mathcal{A}_j\})))$ can learn distinct embedding for $v_a$ and $v_b$, since the composition of three injective functions is still an injective function. If we use $\psi(\cdot)$ to denote $g\circ f$, then it is equivalence to $\psi(\sum_{j\in[1,M]} \beta_j \textsf{SUM}(\{\mathcal{X}(v)| v \in \mathcal{A}_j\}))$, where $\psi(\cdot)$ is an injective function since both $f(\cdot)$ and $g(\cdot)$ are injective. Therefore, there exist some injective functions $\psi(\cdot)$ that allow the AE-aware aggregator to learn distinct node embedding for $v_a$ and $v_b$. Note that since the initial node feature $\mathcal{X}$ is countable and the AE-aware aggregator is injective, the hidden embeddings $\bm{h}^{k-1}(v), k \in[2,K]$ is also countable. Therefore, this argument holds for AE-aware aggregators in all hidden layers as described in (\ref{eq:gae}). Besides, the universal approximation theorem suggest that we can use multi-layer perception (MLP) with at least one hidden layer to approximate any injective function. Therefore, we can use $\textsf{MLP}(\cdot)$ to approximate the injective function $\psi(\cdot)$. As a result, our AE-aware aggregator described in (\ref{eq:gae}) can discriminate the nodes with distinctive Ego-AE feature.

\subsection{Proposition~\ref{prop:match}}

\begin{proof}
As defined in Figure~\ref{fig:gene} and Appendix~\ref{app:subgraph_match}, we have two types of mutations: a) \emph{node mutation} that attaches a new node to a randomly selected node in parent subgraph template; and b) \emph{edge mutation} that randomly adds an edge between two unconnected nodes in parent subgraph template.

Given a graph $G=(\mathcal{V}, \mathcal{E})$, let the matched instance set of a parent subgraph template $S_p=(\mathcal{U}_p, \mathcal{R}_p)$ be $\mathcal{M}_p$. We define the mutated children subgraph template as $S_c=(\mathcal{U}_c, \mathcal{R}_c)$. Based on the definitions of \emph{edge mutation} and \emph{node mutation}, the parent subgraph template is a subgraph of the children subgraph template, \emph{i.e.}, $\mathcal{U}_p \subset \mathcal{U}_c, \mathcal{R}_p \subset \mathcal{R}_c$. Therefore, the matched instances of parent subgraph template will be a partial match of the children subgraph template, \emph{i.e.}, $m_p \subset m_c: \exists m_p \in \mathcal{M}_p, \forall m_c \in \mathcal{M}_c$. Therefore, $m_c$ can be efficiently identified by incrementally extending $m_p$.  
\end{proof}

\begin{algorithm}[t]
	\caption{\underline{GR}aph \underline{A}utormor\underline{P}hic \underline{E}quivalent Network (GRAPE)}
	\label{alg:grape}
	\begin{algorithmic}[1]
		\STATE {Input graph $G = (\mathcal{V}, \mathcal{E})$, node feature $\mathcal{X}(v)$, Ego-AE sets $\{\mathcal{T}_{1}, ..., \mathcal{T}_{L}\}$ for $L$ subgraph templates, layer $k\in[1,K_1]$, non-linearity $\sigma(\cdot)$; }
		\STATE{Node embedding $\bm{h}(v)$, $\bm{h}^0(v) \leftarrow \mathcal{X}(v), \forall \ v\ \in \mathcal{V}$;}
		\STATE{Ground truth $y(v)$; loss function $Loss(\cdot,\cdot)$; epochs $n\in[1,N]$;}
		\FOR{$n \in 1, ..., N$}
		\FOR{$k \in 1, ..., K_1$}
		\STATE \textit{\# AE-aware aggregator with various subgraph templates}\\
		\FOR{$\mathcal{T}_l \in \{\mathcal{T}_{1}, ..., \mathcal{T}_{L}\}$}
		
		\STATE Compute $\bm{h}_l^k(v)$ using (\ref{eq:gae})
		\ENDFOR
		
		\textit{\# Squeeze-and-excitation module to fuse multi templates embeddings}\\
		\STATE compute $\bm{\alpha}^k$ using (\ref{eq:alpha});
		\STATE compute $\bm{h}^k(v)$ using (\ref{eq:fusing});
		\ENDFOR
		\STATE $\hat{y}(v) = MLP(\bm{h}^{K_1}(v))$
		\STATE $Back\_Propagation(Loss(\hat{y}(v), y(v)))$
		\ENDFOR
		\STATE Return $Accuracy(\hat{y}(v), y(v))$
	\end{algorithmic}
\end{algorithm}

\section{More Details for Section~\ref{sec:Method}}

\subsection{GRAPE Algorithm in Section~\ref{sec:gnnarch}}
\label{app:grape}

The GRAPE algorithm~(Algorithm~\ref{alg:grape}) takes a graph $G$, 
node feature vector $\mathcal{X}(v)$ and the Ego-AE sets $\{\mathcal{T}_{1}, ..., \mathcal{T}_{L}\}$ identified by given subgraph templates as input. In each layer, GRAPE uses AE-aware aggregator to transform the features in each Ego-AE set based on  (\ref{eq:gae}). Besides, the embeddings learned from each subgraph template are fused together with a squeeze-and-excitation module based on  (\ref{eq:fusing}) and (\ref{eq:alpha}). Finally, the final layer embedding is transformed by a two-layer multi-layer perception module (MLP) to generate prediction results.

\subsection{Genetic Algorithm in Section~\ref{sec:evoae}}
\label{app:subgraph_match}

\subsubsection{Illustration of Genetic Operations}

\begin{figure}[t]
\centering
\includegraphics[width=0.90\textwidth]{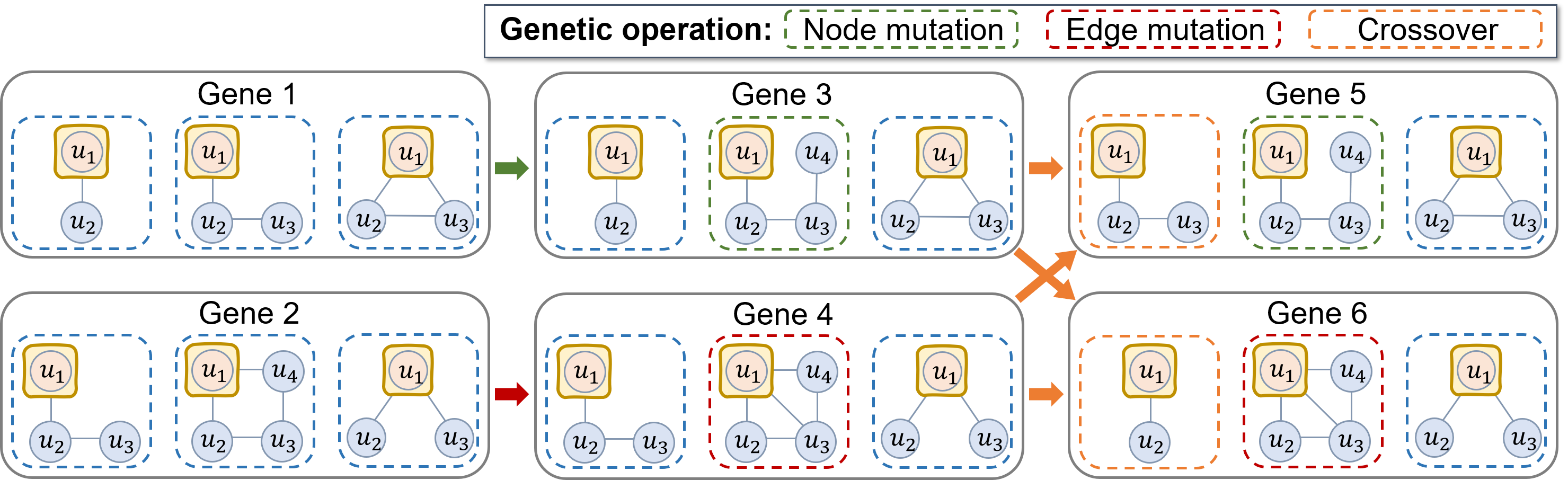}
\caption{Illustration of the \emph{mutate} and \emph{crossover} operations in the proposed genetic algorithm.}
\label{fig:gene}
\end{figure}

Figure~\ref{fig:gene} illustrates the \emph{node mutation}, \emph{edge mutation} and \emph{crossover} operations in the proposed genetic algorithm. Specifically, the \emph{node mutation} will generate a \emph{children subgraph} by randomly adding one node to the input \emph{parent subgraph}, while \emph{edge mutation} generates a \emph{children subgraph} by randomly an edge between two unconnected nodes in the input \emph{parent subgraph}. Besides, the \emph{crossover} operation will randomly exchange some subgraph templates between two genes. These operations effectively allow us to gradually search for slightly more complicated subgraph templates and try out different combinations of subgraph templates.

\subsubsection{Incremental Subgraph Matching}

\begin{figure}[t]
	\centering
	\includegraphics[width=0.9\textwidth]{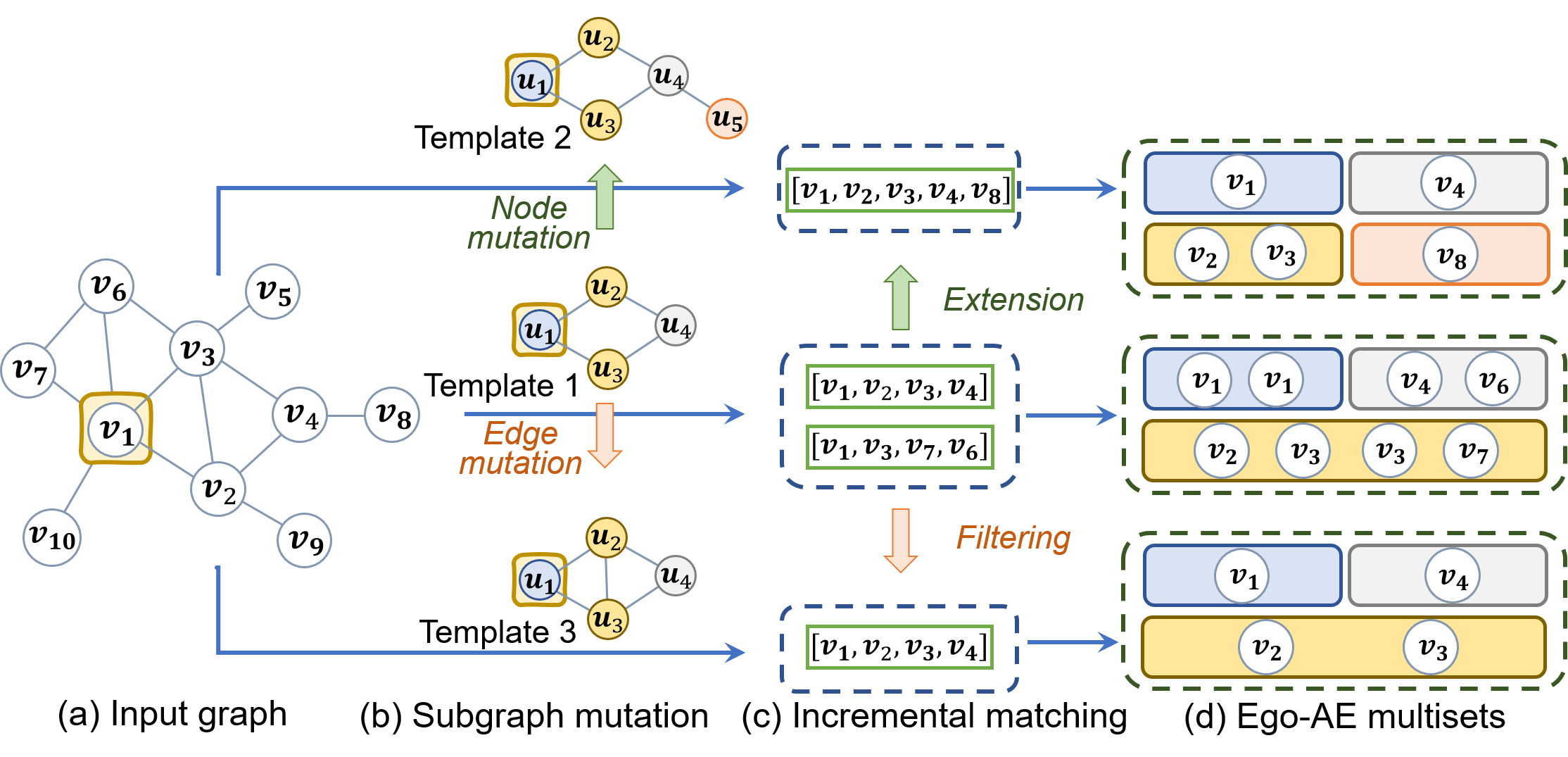}
	\caption{Illustration of identifying Ego-AE sets with incremental subgraph matching.}
	\label{fig:map}
\end{figure} 

To accelerate the matching from the subgraph template to each node's neighborhood, we propose to leverage the similarity between \emph{children subgraph} and \emph{parent subgraph}. Inspired by Proposition~\ref{prop:match}, we design an incremental subgraph matching algorithm that identifies the matched instances of \emph{children subgraph} by only examining the matched instances of \emph{parent subgraph}, which is illustrated in Figure~\ref{fig:map}. Specifically, given the matched instances of \emph{Template 1}, \emph{i.e.}, $[v_1, v_2, v_3, v_4]$ and $[v_1, v_3, v_7, v_6]$, we identify the matched instances of its node mutation children \emph{Template 2} by exploring the neighbors of the current matched instances. Since the newly added node is attached to $u_4$, we will only examine the neighbors of nodes that mapped to $u_4$, \emph{i.e.}, $v_6$ and $v_4$. Therefore, we find $v_8$ as a feasible candidate and identify the matched instance of \emph{Template 2} as $[v_1, v_2, v_3, v_4, v_8]$. As for the edge mutation children \emph{Template 3}, we only need to examine the newly added edge between $u_2$ and $u_3$ in the matched instances. We find $v_2$ and $v_3$ are indeed connected but there is no edge between $v_3$ and $v_7$. Therefore, we identify one matched instance for \emph{Template 3}, \emph{i.e.}, $[v_1, v_2, v_3, v_4]$. Complex subgraph templates with numerous nodes usually result in exponential growth in matching computation time compared to the simpler ones~\cite{cordella2004sub}, but they often have much fewer matched instances. Therefore, by leveraging the feature of genetic search with incremental subgraph matching, we can significantly reduce the computation complexity by only examining the matched instances of parent subgraph instead of starting from scratch.

\subsection{Time Complexity}
\label{app:time}

\textbf{GRAPE model.}
Here, we analyze time complexity of one forward pass of GRAPE model. 
Specifically, suppose we have $L$ subgraph templates, each template has $M$ Ego-AE sets and each set contains $Q$ nodes on average, the overall time complexity of training GRAPE with $K$ layers is $\mathcal{O}(|\mathcal{V}|LMQK)$, 
where $|\mathcal{V}|$ is the number of nodes on graph. 
Empirically, the matched neighbor of each subgraph is a subset of each node's neighborhood, \emph{i.e.}, 
$MQ\leq|\mathcal{N}|$. 
Therefore, GRAPE's time complexity is comparable to the popular MPNNs, \emph{e.g.}, GraphSAGE and GCN, 
which typically have a time complexity of $\mathcal{O}(|\mathcal{V}||\mathcal{N}|K)$.

\textbf{Incremental subgraph matching.}
We assume the \emph{parent subgraph} has $\Pi_e$ ego-centered automorphisms and $|\mathcal{M}_p|$ match instances, and each node has $|\mathcal{N}|$ neighbors on the target graph. Then, the complexity of identifying the match instances after adding node \emph{mutation} is $O(|\mathcal{M}_p||\mathcal{N}|\Pi_e)$, since we only need to examine the possible extensions. Similarly, the complexity of examining adding edge \emph{mutation} is $O(|\mathcal{M}_p|\Pi_e)$. They both have significantly lower the $\mathcal{O}(|\mathcal{V}|!|\mathcal{V}|)$ worst case computation complexity deduced by previous work~\cite{cordella2004sub}. The average case computation complexity can not be analytically estimated unless very restrictive assumptions are made. However, the empirical experiments in Table~\ref{tbl:search} and Figure~\ref{fig:search} demonstrate our proposed incremental search algorithm can significantly outperform baselines on real-world datasets. Moreover, similar sampling approach as in GraphSAGE~\cite{hamilton2017inductive} can be adopted to control the size of match instance set $|\mathcal{M}_p|$, which can ensure the computational footprint of our algorithm is feasible. 

\section{Experiments Details}

\subsection{Experiment Setting and Hyper-parameter}
\label{app:experiment}

Following the setting in previous works~\cite{xu2018powerful}, 
we perform a grid search on the following hyper-parameters: 
1) embedding size $\in\{16, 32\}$; 
2) the dropout rate $\in\{0.3, 0.5\}$; 
3) L2 regularization coefficient $\in\{3\cdot10^{-5}, 5\cdot10^{-5}\}$; 
4) initial learning rate $\in\{0.01, 0.03\}$, which is decayed by 50\% for every 100 epochs. To improve the robustness of experiment results, we report the average and standard deviation of each model's performance over 10 runs. In each run, we randomly split the datasets into 60\% training set, 20\% validation set and 20\% test set. Specifically, we use the training set to learn the models, and report the classification performance o test set. We train each model for 500 epochs with early stopping of 50 window size, 
\emph{i.e.} the training is terminated if the model's performance on validation set does not improve for consecutive 50 epochs. Our model and all the baseline models are implemented in Pytorch~\cite{paszke2019pytorch} with the Adam optimizer. We evaluate them on a single machine with 4  NVIDIA GeoForce RTX 2080 GPUs.

\subsection{Subgraph Templates Design}
\label{app:subgraph}

\begin{figure}[ht]
	\centering
	\includegraphics[width=0.90\textwidth]{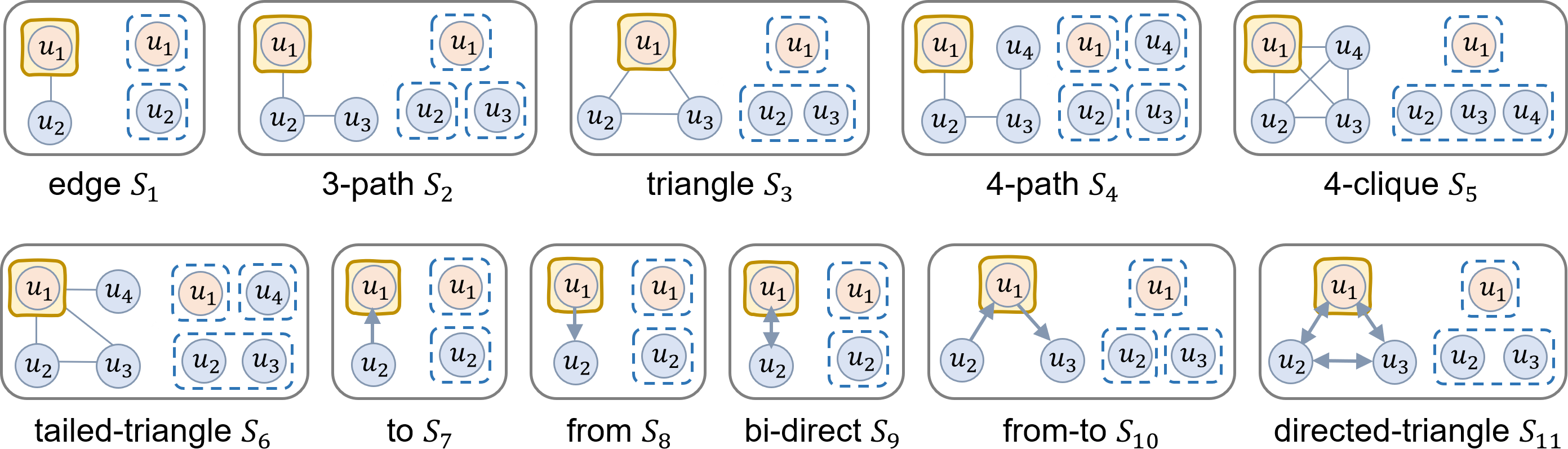}
	\caption{Illustration of the designed subgraph templates.}
	\label{fig:motif}
\end{figure}

We design multiple subgraph templates to allow GRAPE to capture various automorphic equivalences, which are presented in Figure~\ref{fig:motif}. Here, we discuss the motivations for their design.

\begin{itemize}[leftmargin=12px]
\item[1)] \emph{Edge $S_1$}: it captures the basic connection in graph data, 
and has two sets of Ego-AE nodes, \emph{i.e.}, $\{u_1\}$ and $\{u_2\}$. 
Therefore, it leads to two Ego-AE sets: 
one contains the ego node itself (corresponding to $u_1$) and the other contains all the 1-hop neighbors (corresponding to $u_2$).  

\item[2)] 3-path $S_2$: it captures the 2-hop neighborhood of the ego node and partition the neighbors into three Ego-AE sets based on their hops from ego node, 
which maps to $\{u_1\}$, $\{u_2\}$ and $\{u_3\}$, respectively. 
In the context of citation network, it captures the documents that co-cite one document. 
Besides, it captures the individuals that have common friends in social network.

\item[3)] Triangle $S_3$: This template captures an important pattern in graph data, \emph{i.e.}, triangle. 
Based on the triadic closure theory~\cite{huang2014mining,hanneman2005introduction}, 
this template captures the strong ties in social graph, \emph{i.e.}, 
the individuals that form triangle structure tend to have similar feeling about an object. 
This template maps the neighborhood into two Ego-AE sets that corresponds to $\{u_1\}$ and $\{u_2, u_3\}$, respectively. 
But differs from edge template $S_1$, $\{u_2, u_3\}$ 
maps to the neighbors that tend to have stronger influence on the ego node.

\item[4)] 4-path $S_4$: similar to the 3-path template, this template maps the nodes in 3-hop neighborhood into 4 Ego-AE sets based on their hops from ego node. It allows the model to access more far away features.

\item[5)] 4-clique $S_5$: this template captures the closed connected communities in graph data, which tend to exhibit the ``homophily effect''~\cite{mcpherson2001birds}. It maps the neighborhood into two Ego-AE sets that corresponds to $\{u_1\}$ and $\{u_2, u_3, u_4\}$, respectively.

\item[6)] Tailed-triangle $S_6$: on the basis of triangle template $S_3$, this template adds an additional neighbor to the ego node. Therefore, it partitions the neighborhood into three Ego-AE sets, \emph{i.e.}, $\{u_1\}$, $\{u_2, u_3\}$ and $\{u_4\}$, where $\{u_2, u_3\}$ identifies the neighbors with strong ties and $\{u_4\}$ identifies the neighbors connected by simple edge.

\item[7)] To $S_7$: it identifies the neighbors that point to the ego node in directed graph. In the context of e-commerce co-purchase network, $u_2$ maps to items that often lead to the purchase of $u_1$.

\item[8)] From $S_8$: it identifies the neighbors that have directed edges from the ego node. In the context of e-commerce co-purchase network, the purchase of ego node $u_1$ often leads to the purchase of $u_2$.

\item[9)] Bi-direct $S_9$: it identifies the neighbors that are connected to and from the ego node, which are the intersection of the nodes identified by template $S_7$ and $S_8$. Therefore, $u_2$ maps to the items that are frequently co-purchased with $S_{11}$.

\item[10)] From-to $S_{10}$: it maps to the unions of the nodes identified by template $S_7$ and $S_8$. Specifically, it partitions the neighborhood into three Ego-AE sets, \emph{i.e.} $\{u_1\}$, $\{u_2\}$ and $\{u_3\}$, which correspond to the ego node itself, the nodes that point to ego node, and the nodes that are pointed from ego node, respectively.

\item[11)] Directed-triangle $S_{11}$: similar with the triangle template $S_3$, this template captures the triangle patterns in directed graph setting. Specifically, it leads to two Ego-AE sets, which correspond to $\{u_1\}$ and $\{u_2, u_3\}$, respectively.
\end{itemize}

Specifically, we use $\{S_1, S_2, S_3, S_4, S_6\}$ for citation datasets, $\{S_1, S_2, S_3, S_5, S_6\}$ for social datasets, and $\{S_7, S_8, S_9, S_{10}, S_{11}\}$ for amazon dataset.

\subsection{Results with Dummy and Random Initialized Node Feature}
\label{app:dummy}

Here, we present the model performance on datasets with dummy and random initialized node feature in Table~\ref{tbl:dummyresults} and Table~\ref{tbl:randomresults}, 
where the original node feature vectors are replaced by all-ones vectors and randomly generated vectors. 
We can observe that GRAPE consistently outperforms all baseline models with both experiment settings, where the relative accuracy gain over the best baseline models reaches up to 44.7\% and 48.6\%, respectively. 
It indicates GRAPE is expressive with or without node feature, which showcases its capacity in capturing rich structural features.

\begin{table}[ht]
\caption{Classification accuracy on datasets with dummy node feature (\%). The best-performing GNNs are in boldface.}
\centering
\setlength\tabcolsep{5pt}
\begin{tabular}{p{1.7cm}<{\centering}m{1.15cm}<{\centering}m{1.15cm}<{\centering}m{1.15cm}<{\centering}m{1.15cm}<{\centering}m{1.15cm}<{\centering}m{1.15cm}<{\centering}m{1.15cm}<{\centering}m{1.15cm}<{\centering}}
\toprule
& \multicolumn{5}{c} {\textbf{Social}} & \multicolumn{2}{c} {\textbf{Citation}} &  \textbf{Ecomm.} \\ 
\cmidrule(lr){2-6} 
\cmidrule(lr){7-8} 
\cmidrule(lr){9-9}
Model &  Hamilton &  Lehigh &  Rochester& JHU & Amherst &  Cora&  Citeseer&  Amazon\\ 
\midrule

GCN & 19.5$\pm$1.5& 23.4$\pm$1.3& 22.4$\pm$1.6 &19.3$\pm$0.8 & 18.1$\pm$1.7 & 31.0$\pm$1.2 & 21.5$\pm$0.9 & 38.8$\pm$1.0\\

GraphSAGE & 18.8$\pm$3.3 & 20.6$\pm$3.1 & 20.4$\pm$2.3 & 18.6$\pm$2.2 & 17.0$\pm$2.4 & 29.7$\pm$1.6 & 20.0$\pm$0.9 & 38.5$\pm$1.0\\

GIN & 22.7$\pm$5.1& 19.2$\pm$2.5& 22.1$\pm$1.7 &24.2$\pm$4.0 & 18.9$\pm$3.9 & 29.5$\pm$1.3 & 21.1$\pm$1.5 & \underline{39.1$\pm$0.9}\\

GAT & 16.8$\pm$1.4 & 23.5$\pm$3.4 & 21.5$\pm$0.9 & 17.6$\pm$1.1 & 16.9$\pm$2.3 & 25.8$\pm$2.9 & 18.2$\pm$0.9 & 38.7$\pm$1.2\\

Geniepath & \underline{27.5$\pm$2.8} & 23.3$\pm$1.7 & 21.7$\pm$1.5 & 21.4$\pm$3.6 & \underline{25.3$\pm$3.4} & 31.4$\pm$1.7 & 19.1$\pm$1.0 & 38.2$\pm$0.9\\

\midrule
Meta-GNN & 23.7$\pm$1.6 & \underline{25.6$\pm$1.5} & \underline{25.8$\pm$1.1} & \underline{28.8$\pm$2.3} & 23.1$\pm$3.2 & 30.4$\pm$1.4 & \underline{24.5$\pm$1.7} & 38.6$\pm$1.1\\

Mixhop & 19.8$\pm$0.0 & 23.1$\pm$0.1 & 17.9$\pm$0.1 & 18.6$\pm$0.2 & 17.3$\pm$0.1 & \underline{31.9$\pm$0.1} & 18.1$\pm$0.0 & 38.9$\pm$0.1\\

\midrule

DE-GNN & 21.7$\pm$2.1 & 24.7$\pm$2.2 & 18.0$\pm$0.0 & 18.3$\pm$0.1 & 18.6$\pm$2.2 & 31.8$\pm$0.1 & 17.9$\pm$0.3 & 38.9$\pm$0.0\\

\midrule

GRAPE & \textbf{39.8$\pm$3.8} & \textbf{28.9$\pm$2.4} & \textbf{32.5$\pm$1.4} &\textbf{35.8$\pm$2.3} & \textbf{36.6$\pm$3.2} & \textbf{34.4$\pm$3.3} & \textbf{26.3$\pm$0.8} & \textbf{42.9$\pm$0.7} \\
\bottomrule
\end{tabular}
\label{tbl:dummyresults}
\end{table}

\begin{table}[ht]
\caption{Classification accuracy on datasets with random initialized node feature (\%). The best-performing GNNs are in boldface.}
\centering
\setlength\tabcolsep{5pt}
\begin{tabular}{p{1.7cm}<{\centering}m{1.15cm}<{\centering}m{1.15cm}<{\centering}m{1.15cm}<{\centering}m{1.15cm}<{\centering}m{1.15cm}<{\centering}m{1.15cm}<{\centering}m{1.15cm}<{\centering}m{1.15cm}<{\centering}}
\toprule
& \multicolumn{5}{c} {\textbf{Social}} & \multicolumn{2}{c} {\textbf{Citation}} &  \textbf{Ecomm.} \\ 
\cmidrule(lr){2-6} 
\cmidrule(lr){7-8} 
\cmidrule(lr){9-9}
Model &  Hamilton &  Lehigh &  Rochester& JHU & Amherst &  Cora&  Citeseer&  Amazon\\ 
\midrule

GCN & 19.7$\pm$3.3 & 23.2$\pm$0.1 & 21.9$\pm$0.8 &19.0$\pm$2.4 & 17.2$\pm$1.2 & \underline{39.9$\pm$9.8} & \underline{29.6$\pm$9.6} & \underline{38.9$\pm$0.1}\\

GraphSAGE & 17.3$\pm$5.6 & 15.2$\pm$8.2 & 17.8$\pm$6.2 & 15.4$\pm$7.9 & 15.0$\pm$5.9 & 34.9$\pm$0.7 & 23.0$\pm$2.4 & 35.9$\pm$1.1\\

GIN & \underline{35.7$\pm$2.4} & 24.3$\pm$2.1 & \underline{32.1$\pm$1.6} & \underline{32.4$\pm$3.0} & \underline{30.0$\pm$5.4} & 27.3$\pm$0.1 & 20.0$\pm$0.0 & 37.0$\pm$0.2\\

GAT & 17.6$\pm$3.4 & 24.3$\pm$2.1 & 21.9$\pm$1.8 & 18.0$\pm$3.4 & 16.2$\pm$1.4 & 25.1$\pm$2.2 & 19.2$\pm$1.7 & 38.6$\pm$0.6\\

Geniepath & 28.1$\pm$3.1 & 23.3$\pm$2.1 & 21.4$\pm$0.6 & 22.1$\pm$4.3 & 24.9$\pm$3.6 & 31.2$\pm$7.2 & 18.0$\pm$3.5 & 38.0$\pm$1.2\\

\midrule
Meta-GNN & 24.7$\pm$2.6 & \underline{25.1$\pm$1.8} & 26.0$\pm$2.2 & 28.9$\pm$3.1 & 23.6$\pm$2.8 & 30.1$\pm$1.6 & 25.3$\pm$0.7 & 38.4$\pm$0.7\\

Mixhop & 21.3$\pm$0.2 & 24.1$\pm$1.3 & 18.4$\pm$0.3 & 17.6$\pm$0.4 & 18.5$\pm$0.9 & 31.8$\pm$0.5 & 19.8$\pm$1.2 & 38.7$\pm$0.4\\

\midrule

DE-GNN & 21.9$\pm$1.6 & 23.2$\pm$0.1 & 18.0$\pm$0.0 & 21.6$\pm$1.0 & 20.7$\pm$3.3 & 31.9$\pm$0.0 & 18.2$\pm$0.1 & \underline{38.9$\pm$0.0}\\

\midrule

GRAPE & \textbf{38.9$\pm$3.0} & \textbf{31.1$\pm$1.5} & \textbf{34.3$\pm$0.6} &\textbf{34.0$\pm$2.0} & \textbf{37.2$\pm$3.5} & \textbf{40.7$\pm$3.5} & \textbf{44.0$\pm$4.7} & \textbf{39.1$\pm$3.5} \\
\bottomrule
\end{tabular}
\label{tbl:randomresults}
\end{table}

\subsection{Results with Different Subgraph Templates}
\label{app:result_subgraph}

\begin{table}[ht]
\caption{Classification accuracy with different subgraph templates on \emph{Lehigh} dataset (\%).}
\centering
\setlength\tabcolsep{5pt}
\begin{tabular}{p{1.7cm}<{\centering}m{1.15cm}<{\centering}m{1.15cm}<{\centering}m{1.15cm}<{\centering}m{1.15cm}<{\centering}m{1.15cm}<{\centering}m{1.15cm}<{\centering}m{1.15cm}<{\centering}m{1.15cm}<{\centering}}
\toprule
& \multicolumn{6}{c} {\textbf{Subgraph Templates}} \\ 
\cmidrule(lr){2-7} 
Model &  $S1$ & $S2$ & $S3$ & $S4$ & $S5$ &  $S6$\\ 
\midrule

GRAPE & 23.3$\pm$1.7 & 22.6$\pm$1.3 & 26.1$\pm$4.0 & 22.9$\pm$0.6 & 23.2$\pm$1.3 & 23.3$\pm$0.9 \\

\bottomrule
\end{tabular}
\label{tbl:subgraphs}
\end{table}

\subsection{License of Assets}
The source code will be shared under MIT license. All the datasets used in this research is public available. 
\end{document}